\documentclass[mnsc,nonblindrev]{informs3}
\RequirePackage{tgtermes}
\RequirePackage{newtxtext}
\RequirePackage{newtxmath}
\RequirePackage{bm}
\RequirePackage{endnotes}
\OneAndAHalfSpacedXI
\usepackage{appendix}
\usepackage[utf8]{inputenc} % allow utf-8 input
\usepackage[T1]{fontenc}    % use 8-bit T1 fonts
\usepackage{hyperref}       % hyperlinks
\usepackage{url}            % simple URL typesetting
\usepackage{booktabs}       % professional-quality tables
\usepackage{xcolor}
\usepackage{amsmath}
\usepackage{amssymb}
\usepackage{amsfonts}       % blackboard math symbols
\usepackage{nicefrac}       % compact symbols for 1/2, etc.
\usepackage{microtype}      % microtypography
\usepackage{xcolor}         % colors
\usepackage{enumitem}
\usepackage{wrapfig}
\definecolor{framecolor}{rgb}{0, 0.502, 0.502}
\definecolor{bgcolor}{rgb}{0.960, 0.984, 1.000}
\usepackage{multirow}
\usepackage{array}
\usepackage{siunitx} 
\usepackage{longtable} 
\usepackage{tcolorbox}
\newtcolorbox{mybox}[1]{colback=bgcolor,colframe=framecolor,fonttitle=\bfseries,title=#1}
\usepackage{graphicx}
\usepackage{caption}
\usepackage{subcaption}
\usepackage{wrapfig}
\usepackage{multirow}
\usepackage{algorithm}
\usepackage{algpseudocode}
\usepackage{tikz}
\usepackage{threeparttable}
\usepackage{algpseudocode}
\usepackage{color,soul}
\usepackage{xr-hyper}
\usepackage{hyperref}
\usepackage{graphicx} 
\algnewcommand{\algorithmicand}{\textbf{ and }}
\algnewcommand{\algorithmicor}{\textbf{ or }}
\algnewcommand{\OR}{\algorithmicor}
\algnewcommand{\AND}{\algorithmicand}
\algrenewcommand{\algorithmiccomment}[1]{\hfill// #1}
 \def\bibfont{\small}%
\EquationsNumberedThrough    % Default: (1), (2), ...
\TheoremsNumberedThrough     % Preferred (Theorem 1, Lemma 1, Theorem 2)
\ECRepeatTheorems

\MANUSCRIPTNO{IJDS-0001-2024.00}

\makeatletter
\newcommand*{\addFileDependency}[1]{% argument=file name and extension
  \typeout{(#1)}
  \@addtofilelist{#1}
  \IfFileExists{#1}{}{\typeout{No file #1.}}
}

\makeatother
\algrenewcommand\algorithmicrequire{\textbf{Input:}}
\algrenewcommand\algorithmicensure{\textbf{Output:}}
\algnewcommand{\LineComment}[1]{\Statex \(\triangleright\) #1}
\begin{document}
\RUNAUTHOR{Lin et al.} 
\RUNTITLE{Causal Discovery Should Embrace the Wisdom of the Crowd}
\TITLE{Causal Discovery Should Embrace the Wisdom of the Crowd}

\ARTICLEAUTHORS{%
\AUTHOR{\normalsize Ryan Feng Lin\textsuperscript{a}\thanks{Equal contribution.}, Yuantao Wei\textsuperscript{a}\footnotemark[1], Huiling Liao\textsuperscript{b}, Xiaoning Qian\textsuperscript{c}, Shuai Huang\textsuperscript{a}\thanks{Corresponding author.}}
\AFF{\textsuperscript{a}  Department of Industrial and
Systems Engineering, University of Washington, Seattle, WA 98195, USA\\
\textsuperscript{b} Department of Applied Mathematics, Illinois Institute of Technology\\
\textsuperscript{c}Department of Electrical \& Computer Engineering, Texas A\&M University\\
\EMAIL{\{ryanflin, yuantw2, shuaih\}@uw.edu, hliao13@illinoistech.edu, xqian@ece.tamu.edu}} %, \URL{}}
% Enter all authors
} % end of the block

% Causal discovery lies at the heart of scientific understanding, yet remains a computationally intractable challenge due to its NP-hard nature. Existing algorithmic approaches rely solely on observational data and centralized learning, which struggle to scale due to the super-exponential search space and inherent ambiguities in causal structure identification. This position paper proposes a fundamental shift: re-framing causal learning as a distributed, collective intelligence problem, wherein diverse human contributors, each holding partial, noisy, but meaningful knowledge, collaboratively construct a global causal model. 
\ABSTRACT{
\small This position paper argues for recognizing an emerging paradigm of \textit{causal learning by wisdom of the crowd}. Recent developments in government, industry, and research point to the rise of decentralized and crowd-based approaches within causal modeling, where causal knowledge distributed across many contributors can be systematically elicited and integrated with causal learning workflows. In this paradigm, causal learning becomes a distributed decision-making problem: each participant contributes partial and potentially noisy knowledge, while collective contributions help construct a global causal structure. This direction is enabled by advances in crowdsourcing platforms, expert knowledge elicitation, aggregation techniques, and large language model (LLM)-augmented information acquisition. Its promise is increasingly visible in early research and emerging real-world practices. Building on this momentum, we outline a framework for crowd-based causal learning spanning elicitation, modeling, aggregation, and optimization. We further discuss the opportunities and challenges introduced by this paradigm and call for interdisciplinary collaboration across causal learning, collective intelligence, human-AI interaction, and decision science.}

% , and discuss how such systems can support both structure learning and causal inference. 

% Through illustrative examples and integration with existing literature, we identify the analytic and algorithmic building blocks required to formalize this paradigm, and articulate open challenges. 

\KEYWORDS{causal learning, wisdom of the crowd, human knowledge, directed acyclic graphs}  

\maketitle
\vspace{-1cm}

\section{Introduction}
\label{intro}

% \textcolor{red}{Shuai's comment: start with the motivation, the existing practices - use our rebuttal material as the first paragraph. Then the next paragraph talks about the novel research questions that arise from these emerging practices}

Emerging practices in government, industry, and research sectors suggest the rise of a new decentralized and crowd-based mode of causal discovery. Examples include the IARPA-funded Bayesian ARgumentation via Delphi (BARD) project \cite{nyberg2022bard, korb2020individuals} and Microsoft Research's Distributed Causal Inference (DCI) effort \cite{microsoft_dci_2017}. Although these efforts differ in focus, i.e., from structured expert elicitation and argumentation to scalable infrastructure for causal inference, they point to a broader and general pattern: causal knowledge is being produced through combining the wisdom of a crowd. We believe such practices should not be viewed as fragmented curiosities, but as signs of an emerging research direction whose intellectual and computational foundations deserve explicit study.

As a central branch of causal research, causal discovery seeks to infer causal structures from data \cite{pearl2009causality, sucar2025causal}. Existing methods remain essential, but their theory and practice also show why observational data alone is often insufficient. Causal structures may not be identifiable without additional assumptions or auxiliary information, such as interventions or domain knowledge \cite{carey2025distinguishability}, especially in large-scale systems with many interacting, unobserved, or poorly measured factors \cite{kitson2025causal}. Interventions reduce ambiguity, but they are often costly, constrained, or infeasible \cite{glymour2019review,murphy2003optimal}. These challenges underscore the value of expert knowledge as a complementary information source for causal discovery \cite{cano2011method, gururaghavendran2025can, heckerman1995learning}.

Domain experts may know that some relations are plausible, for example, that transcription factors regulate the expression of their target genes, or that others are mechanistically, temporally, or contextually implausible, thereby reducing the search space \cite{constantinou2023impact}. Traditional approaches incorporate such knowledge through priors over graph structures or constraints that forbid or enforce edges \cite{borboudakis2012incorporating, hasan2022kcrl, gonzales2022hybrid}. Yet emerging causal practices raise questions not fully addressed by this framing. Expert knowledge is often compressed into hard constraints or heuristic rules, obscuring uncertainty, context, and disagreement. It is also commonly drawn from a small number of experts, despite the incompleteness and bias of any single perspective. In complex domains such as biomedicine, economics, engineering, and policy, causal knowledge is instead fragmented across heterogeneous contributors. This motivates a central question: \textit{Can we systematically harness distributed causal knowledge from a crowd of contributors, each holding a partial piece of the puzzle, to construct a reliable causal representation among variables of interest?}

\textbf{Our Position.} We argue that \textbf{the time is ripe to recognize ``causal learning by wisdom of the crowd'' as an emerging and complementary research paradigm}. It is not intended to replace score-based \cite{yuan2011learning, ng2024score}, constraint-based \cite{spirtes2000pcfci, li2019constraint}, differentiable \cite{huang2012sparse, zheng2018dags, ban2024differentiable}, interventional \cite{varici2024interventional, li2023causal}, or other data-driven causal discovery methods. Rather, it addresses a distinct layer of causal practice: how distributed causal beliefs can be elicited, modeled, denoised, reconciled, aggregated, and integrated into existing causal learning pipelines. We envision a \textit{decentralized intelligence solicitation platform} that treats domain experts, non-domain experts, and possibly simulated agents such as Large Language Models (LLMs) as distributed ``puzzle-solvers,'' each contributing partial and noisy insights into local causal relationships. Through principled elicitation and aggregation, these fragments may be assembled into a coherent global causal representation unavailable to any single participant. This idea draws inspiration from crowdsourcing discovery \cite{vaughan2018making, thompson2020crowdsourcing} and collaborative knowledge systems like Wikipedia \cite{azizifard2022wiki}. It also draws on the ``wisdom of the crowd'' principle \cite{surowiecki2005wisdom}, where diverse and independent judgments can outperform individual judgments as idiosyncratic errors cancel through aggregation \cite{saha2021wisdom}. For causal discovery, however, what is aggregated is not merely a vote, label, or opinion, but causal knowledge: relationships, directional influences, and constraints among variables.

\section{The Emerging Form of Crowd-Based Causal Learning}

\begin{figure}[!b]
    \centering
    \vspace{-1em}
    \includegraphics[width=0.8\linewidth]{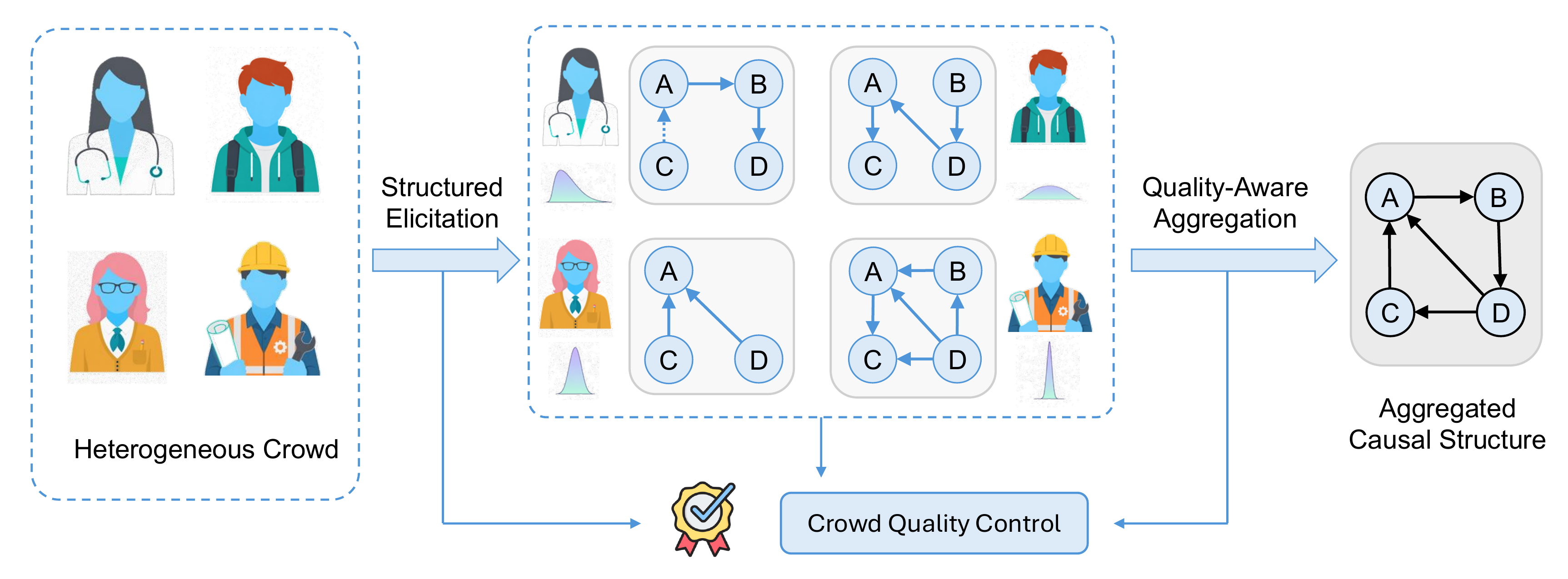}
    \caption{An overview of the crowd-based causal learning.}
    \label{fig:overview}
    \vspace{-0.5em}
\end{figure}

Recent work by \cite{berenberg2018efficient} demonstrates that a large-scale causal network with 394 nodes can be constructed by eliciting and aggregating causal judgments from roughly 100 participants. This study exemplifies an emerging class of crowd-based causal learning systems, whose generic framework is illustrated in Figure \ref{fig:overview}. Unlike traditional approaches treating contributors as homogeneous knowledge sources, these systems explicitly model heterogeneity in individual expertise and behavior, a phenomenon further illustrated in our real-world study (Section \ref{sec:real_exp}). Building on this heterogeneity, the paradigm is two-stage: eliciting causal knowledge and aggregating it into a coherent crowd-level representation (Section \ref{exp_ml}). Across all stages, crowd quality control serves as an underpinning tool. (Section \ref{sec:QC}).

This formulation already appears across several research directions. Collaborative causal learning \cite{kawamata2024collaborative, qiao2023collaborative, addanki2021collaborative} and federated causal learning \cite{tregubov2024tool, li2024federated}, for example, frame causal discovery as a multi-agent process in which distributed contributors jointly construct global causal models through shared data, local models, or partial knowledge. Related ideas are also emerging in practice through systems such as DCI initiative \cite{microsoft_dci_2017} and the BARD platform \cite{nyberg2022bard, korb2020individuals}. Collectively, these efforts point toward a broader paradigm in which causal knowledge is elicited, refined, and integrated from many distributed contributors rather than a single centralized source.

\begin{table*}[!t]
\centering
\small
\caption{Evolution of expert involvement in causal discovery.}
\begin{tabular}{|l|l|l|}
\hline
\textbf{Phase} & \textbf{Methodology} & \textbf{Role of Expert} \\
\hline
Early Manual Construction & Direct Elicitation & Sole source of structure \\
\hline
Pure Data-Driven Learning & Independence Tests / Scoring & Validation only \\
\hline
Hybrid Learning & Priors / Hard Constraints & Guiding the search space \\
\hline
Distributed Wisdom of the Crowd & Decentralized Solicitation & Distributed ``Puzzle Solving'' \\
\hline
\end{tabular}
% \vspace{-1.5em}
\label{tab:expert_roles}
\end{table*} 

More fundamentally, these developments suggest a shift in the role of expert knowledge in causal learning. As summarized in Table \ref{tab:expert_roles}, causal discovery has followed a centralized paradigm: a single learner integrates observational data while expert knowledge enters as trusted priors or constraints. However, real-world causal knowledge is rarely centralized; it is fragmented across individuals, contexts, and forms of expertise. Emerging research and practice increasingly recognize this distributed structure and seek principled ways to leverage it within decentralized causal learning systems.

\textbf{Driving forces.} The paradigm of crowd-based causal learning is not merely speculative, but responds to practical needs for harnessing distributed causal knowledge. (1) \textbf{More complex systems.} Real-world systems are increasingly large, complex and dynamic \cite{karwowski2025grand, liu2007complexity}. Thus, no single expert is likely to possess complete causal knowledge of all relevant mechanisms, which necessitates the combination of partial expertise from multiple contributors \cite{aminpour2020wisdom, zellner2023exploring}. (2) \textbf{Inaccessible data.} Data-driven causal discovery is often constrained by the cost or infeasibility of data collection, especially in dynamic environments where new data cannot be gathered in time \cite{squires2023causal, lagemann2023deep}. \textbf{(3) Pluralistic knowledge sources.} Many domains accumulate various knowledge sources such as expert experience, prior models, and heuristics \cite{zhang2026guiding, garrido2022integrating}, which remain underused centralized causal learning. (4) \textbf{Social legitimacy for practical deployment.} Acceptance of causal models in industry decision-making is still evolving. Broader deployment requires reconciling context-dependent, sometimes conflicting causal assumptions so that domain experts and stakeholders build shared understanding \cite{quimby2023participatory}.

\textbf{Socio-Technological Enablers.} Recent societal and technological advances have matured to the point where crowd-based causal learning is feasible at scale. We highlight several enablers that make crowd-based causal learning timely and practical: 
(1) \textbf{Crowd-based knowledge infrastructure.} Crowdsourcing platforms like Amazon Mechanical Turk \cite{paolacci2010running} and Zooniverse \cite{simpson2014zooniverse} now provide scalable participation infrastructures that recruit, motivate, and coordinate millions of diverse contributors through structured knowledge elicitation workflows. (2) \textbf{Cognitive and behavioral sciences.} Advances in areas such as causal cognition \cite{Sloman2005Causal, gerstenberg2024counterfactual, felin2024theory}, knowledge and expertise \cite{cooke2014modeling}, or metacognition \cite{o2024modeling, o2022measuring} have deepened our understanding of heterogeneous contributors in the crowd for causal discovery. (3) \textbf{Aggregation mechanisms.} These models or tools have greatly advanced in the past decades, such as Dawid-Skene \cite{tamura2024influence}, GLAD \cite{whitehill2009whose} or HLM \cite{wulearning}. They contribute many ways to assess individual knowledge quality and bias, and yield consensus-level judgments critical for building reliable causal graphs from distributed sources.
(4) \textbf{LLMs} like GPT-5 \cite{leon2025gpt} offer more interactive and user-friendly interfaces for knowledge elicitation, while providing scalable knowledge amplification by simulating experts and generating structured causal hypotheses to complement human input \cite{alaa2024large}.
(5) \textbf{Human-AI collaboration tools.} Advances in human-computer interaction (HCI), knowledge engineering and hybrid intelligence systems lower the barriers for both domain experts and laypersons to contribute causal insights via interactive interfaces \cite{zhong2024ai, kale2021causal} and validate structured knowledge (e.g., causal pathways, diagrams) \cite{kiciman2022causal}. (6) \textbf{Interdisciplinary Collaboration.} The growing emphasis on interdisciplinary collaboration provides a cultural foundation for causal learning systems to involve not only statisticians and machine learning researchers, but also domain experts, practitioners, social scientists, and stakeholders in the joint construction, validation and deployment of causal models. 

Taken together, crowd-based causal learning links causal discovery to social choice theory \cite{suksompong2024expanding, kelly2013social}, truth discovery \cite{shabat2022empirical, li2016survey}, design of experiments \cite{xiao2018optimal, zhang2023active}, crowdsourced labeling \cite{yin2021learning, sheng2019machine}, knowledge fusion \cite{zhao2020multi}, and human-in-the-loop learning \cite{mosqueira2023human, wu2022survey}. This  interdisciplinary convergence raises distinct challenges. Addressing them requires new formulations for expert modeling, incentive design, denoising, aggregation, graph reconciliation, and validation, opening a research frontier between collective intelligence \cite{park2025decision, riedl2021quantifying} and structural causal modeling \cite{feuerriegel2024causal, pearl2018book}.

\section{Characterizing the Crowd: Heterogeneity and Reliability}
\label{sec:real_exp}
% \subsection{A Thought Experiment}

\begin{wrapfigure}{r}{0.52\textwidth}
\centering
% \vspace{-5mm}
\includegraphics[width=0.4\textwidth]{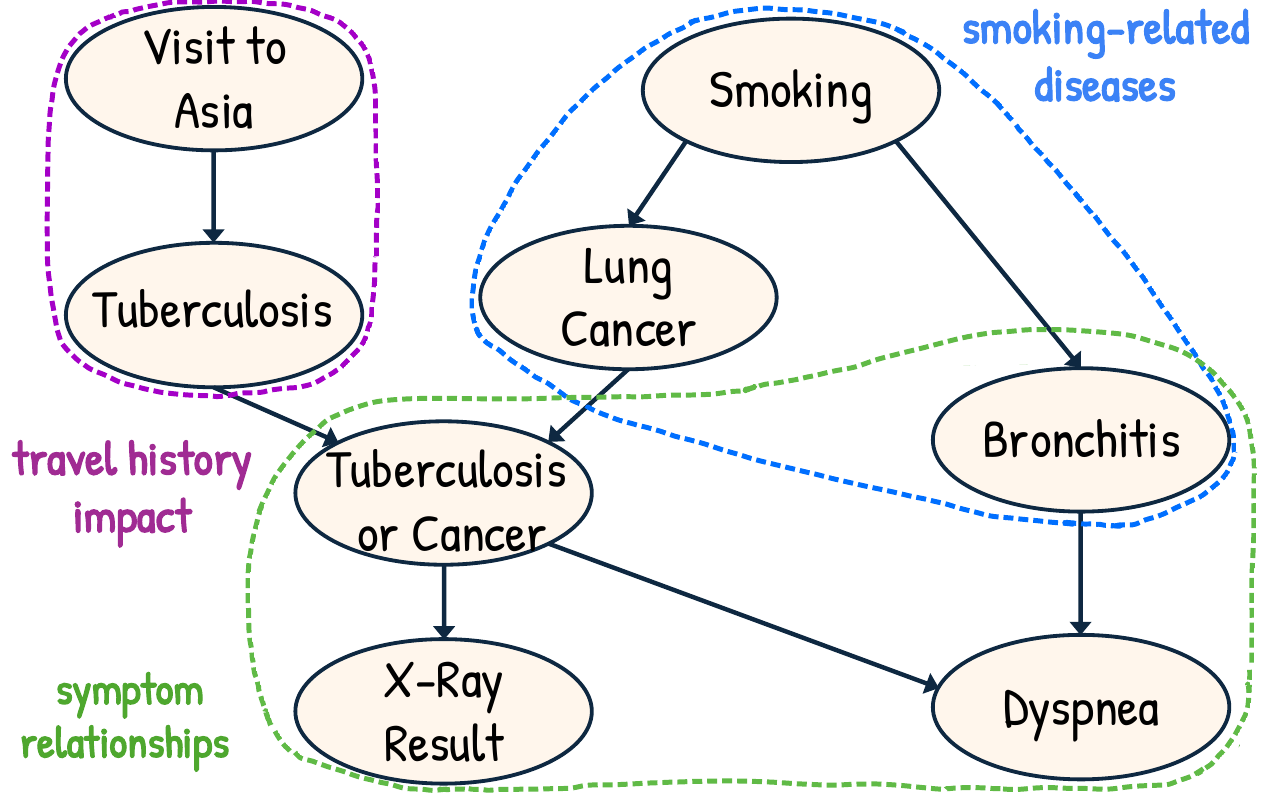}
\caption{Bayesian network \textit{Asia} with eight nodes and eight causal links representing a simplified respiratory diagnosis scenario. }% The links include travel history impact (purple), smoking-related diseases (blue), and symptom relationships (green).}
\label{fig0}
\end{wrapfigure}

We now take a closer look at the core of crowd-based causal learning: the crowd. The individual contributors who make up the crowd are not interchangeable sources of causal judgments, since they may differ in many aspects. Thus, explicitly defining such heterogeneity is necessary before eliciting and safely aggregating their causal wisdom. Bayesian networks (BNs), as natural Directed Acyclic Graph (DAG) representations of causal structures, provide a suitable testbed for illustrating this issue. We start with \textit{Asia}, a benchmark BN (see Figure \ref{fig0}), to perform a thought experiment. Imagine two experts: \textbf{Expert A} is a pulmonologist who knows smoking-related diseases (blue) and symptom relationships (green) very well, but has no knowledge about travel history impact (purple) and will leave the link \textit{Visit to Asia}\textrightarrow \textit{Tuberculosis} unspecified. \textbf{Expert B} is a practitioner with broad but uncertain knowledge across variables. They may incorrectly believe \textit{Tuberculosis or Cancer} causes \textit{Bronchitis}, or be unsure of the causality between \textit{Smoking} and \textit{Lung Cancer}. In decision-making, Expert A's input can be trusted where available, even with some edges uncovered, while the less accurate input of Expert B should be taken with a grain of salt everywhere despite broader coverage, which necessitates a unified framework characterizing diverse expert types.  

\subsection{A Real-World Case Study}
\label{real_exp}

Guided by the above thought experiment, we conducted a proof-of-concept study with the same \textit{Asia} network as the main testbed to understand better the plausibility, potentials and pitfalls of our envisioned paradigm of causal learning by wisdom of the crowd. In this study, 20 participants were recruited as expert informants to share their causal knowledge about the \textit{Asia} network through a structured survey. Participants were not given access to the ground-truth network. Instead, they were provided with detailed textual descriptions of variables in each network, and were asked to assess pairwise causal relationships among the variables. This study followed Institutional Review Board (IRB) procedures, as detailed in Appendix \ref{irbs}. Survey design and experimental procedures are described in Appendix \ref{survey}, and additional network data can be found in Appendix \ref{bnbench}. Figure \ref{fig:complexity_two_panel} visualizes the absolute value distributions of causal scores given by the participants, offering an immediate impression that expert causal knowledge is far from uniform at both the participant and query levels. In the left panel, markedly different violin shapes across participants reveal strong heterogeneity: some concentrate near extreme values indicating consistently high confidence, while others show more variable judgments across queries through their broad, flat distributions. On the right panel, queries likewise induce diverse distributions across the participants. Some queries exhibit more ambiguity and disagreement, whereas others reach broad consensus. Together, these patterns highlight the complex, high-dimensional nature of expert causal knowledge.

\begin{figure}[!h]
  \centering
  \begin{subfigure}[c]{0.43\textwidth}
    \centering
    \includegraphics[width=\linewidth]{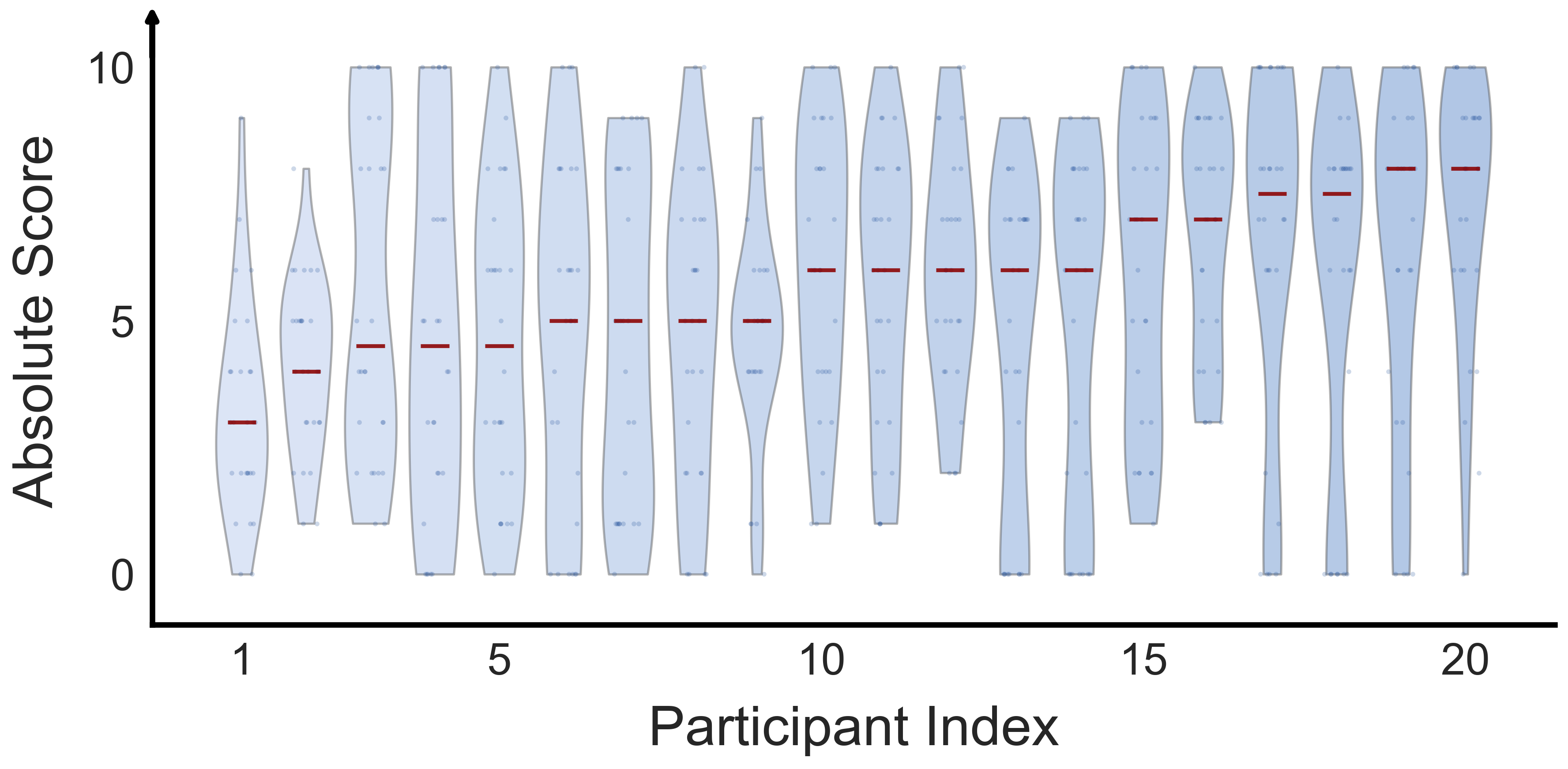}
    \caption{Expert level}
    \label{fig:complexity_expert}
  \end{subfigure}
  \hfill
  \begin{subfigure}[c]{0.43\textwidth}
    \centering
    \includegraphics[width=\linewidth]{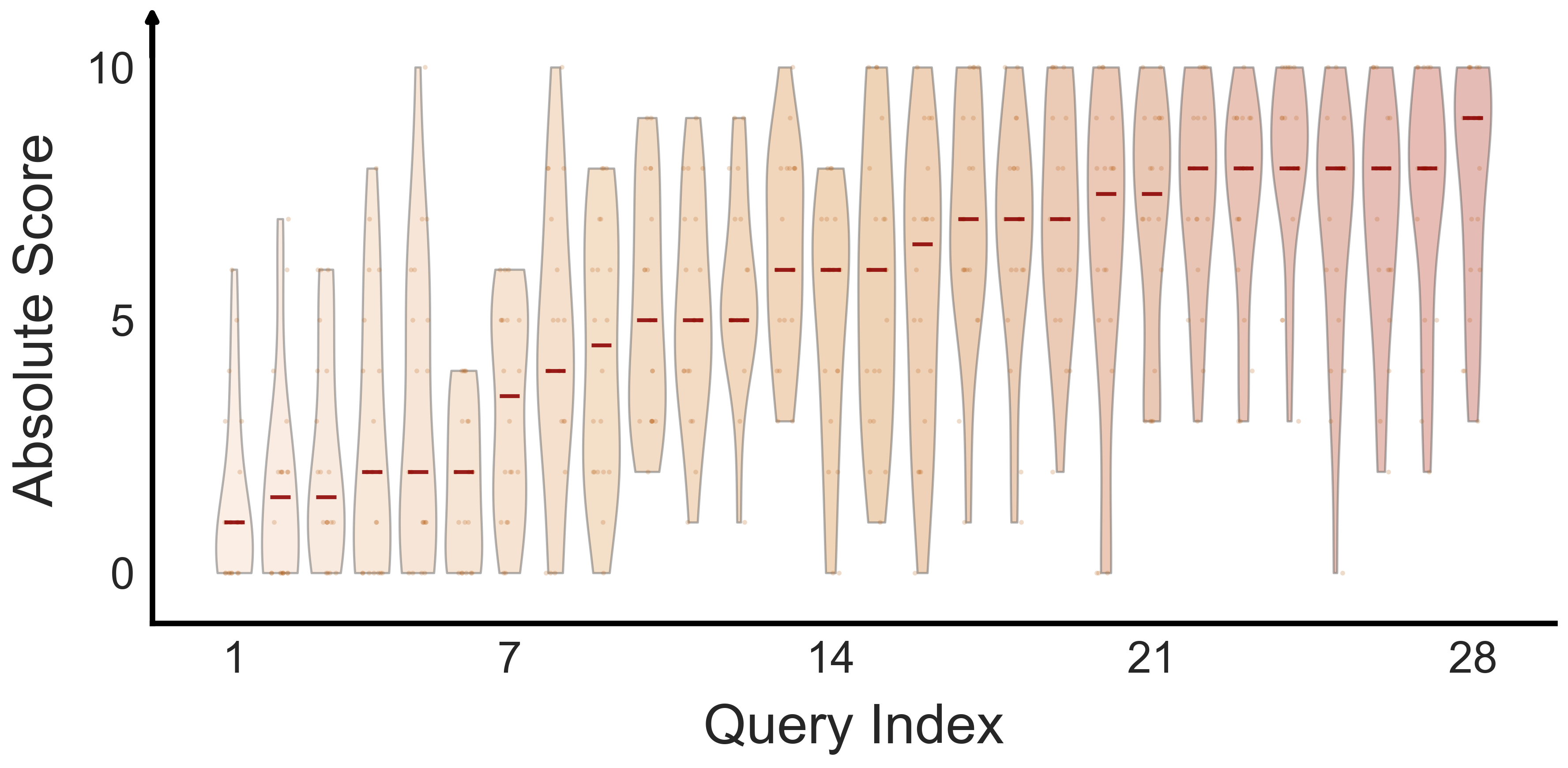}
    \caption{Query level}
    \label{fig:complexity_query}
  \end{subfigure}
  \caption{Expert- and query-level knowledge complexity sorted by median absolute score.}
  \label{fig:complexity_two_panel}
\end{figure}

\subsection{Insights of the Real-World Study} \label{rwinsights}

\begin{wrapfigure}{r}{0.45\textwidth}
    \centering
    \includegraphics[width=1\linewidth]{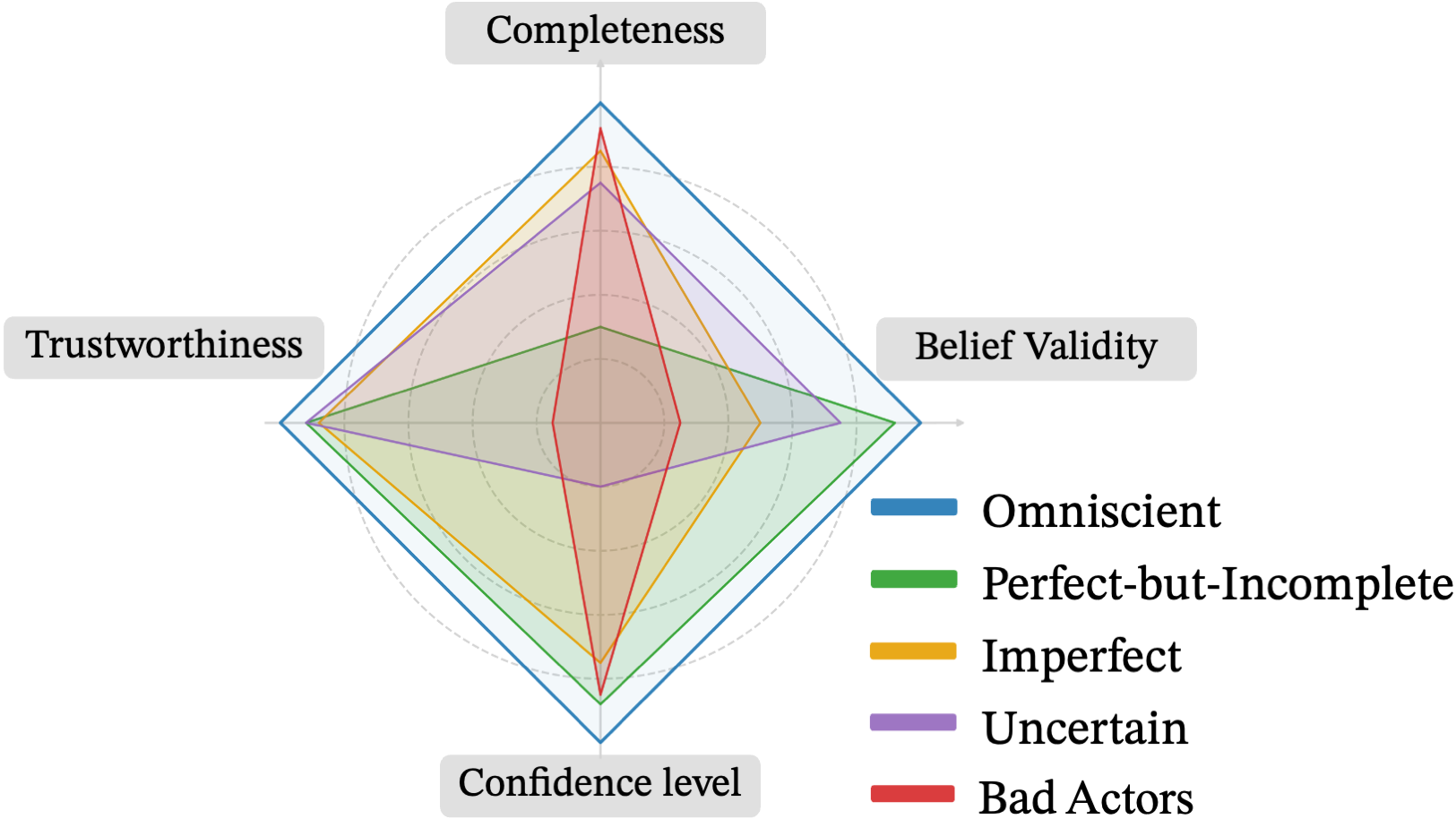}
    \caption{Illustration of expert types in four dimensions.}
    \label{fig:experts_radar}
    \vspace{-0.2em}
\end{wrapfigure}

\textbf{Characterizations of Knowledge Quality.} Experimental evidence reveals that the reliability of expert knowledge can vary widely. An expert might possess highly accurate knowledge about certain causal links (edges), while remaining uncertain or completely ignorant about others. % For any given edge in the Bayesian network, an expert could either (a) know the relationship with near certainty (and likely be correct), (b) have only a rough or biased guess, or (c) have no knowledge at all. 
This heterogeneity demands a conceptual framework to categorize experts by the scope and accuracy of their knowledge. To this end, we recommend a taxonomy grounded in four key dimensions: 
% \begin{enumerate}[topsep=0pt,itemsep=-1ex,partopsep=1ex,parsep=1ex, leftmargin=*]
% \item[(1)] 
\textbf{{(1) Completeness}} reflects the breadth of an expert's causal knowledge, which determines whether it is feasible to obtain fully informative knowledge from the expert. When knowledge is incomplete, they may only cover a subset of causal relationships, potentially leading to partial, myopic or even inconsistent inputs.
% \item[(2)] 
\textbf{{(2) Belief validity}} captures how well an expert's beliefs align with the underlying ground truth, which can be measured by the performance evaluation metrics such as specificity or accuarcy. Within an expert's specialization areas, they tend to offer more accurate and interpretable causal assessments, whereas reliability declines significantly with their beliefs becoming biased or erroneous outside these areas. 
% \item[(3)] 
\textbf{{(3) Confidence level}} is not necessarily associated with the validity of their beliefs, but shapes how cautious or assertive their behaviors are. Beyond correctness, experts differ in their confidence about what they know. %Some express appropriate uncertainty when unsure, while others may be overconfident. 
% \item[(4)] 
\textbf{{(4) Trustworthiness}} captures the presence of bad or adversarial behaviors, as not all experts make good-faith efforts to contribute knowledge to ensure the reliability. Some experts behave strategically, maliciously, or carelessly, thereby introducing significant noise or deliberate misinformation. % Thus, recognizing such uncooperative individuals is essential to ensure the reliability of expert knowledge.
% \end{enumerate}
%The following categories are conceptual extremes: real-world experts may lie on a spectrum between them. 

% \begin{figure}[!t]
%     \centering
%     \includegraphics[width=1\linewidth]{figs/recall_efficiency.png}
%     \caption{A comparison of cost-effectiveness between expert knowledge and observational data for causal learning. \textcolor{red}{adjust based on my writeup}}
%    \label{fig:cost_effectiveness}
% \end{figure}

\begin{figure*}[!h]
\centering
\begin{minipage}[t]{0.16\textwidth}
  \centering
  \includegraphics[width=0.6\linewidth]{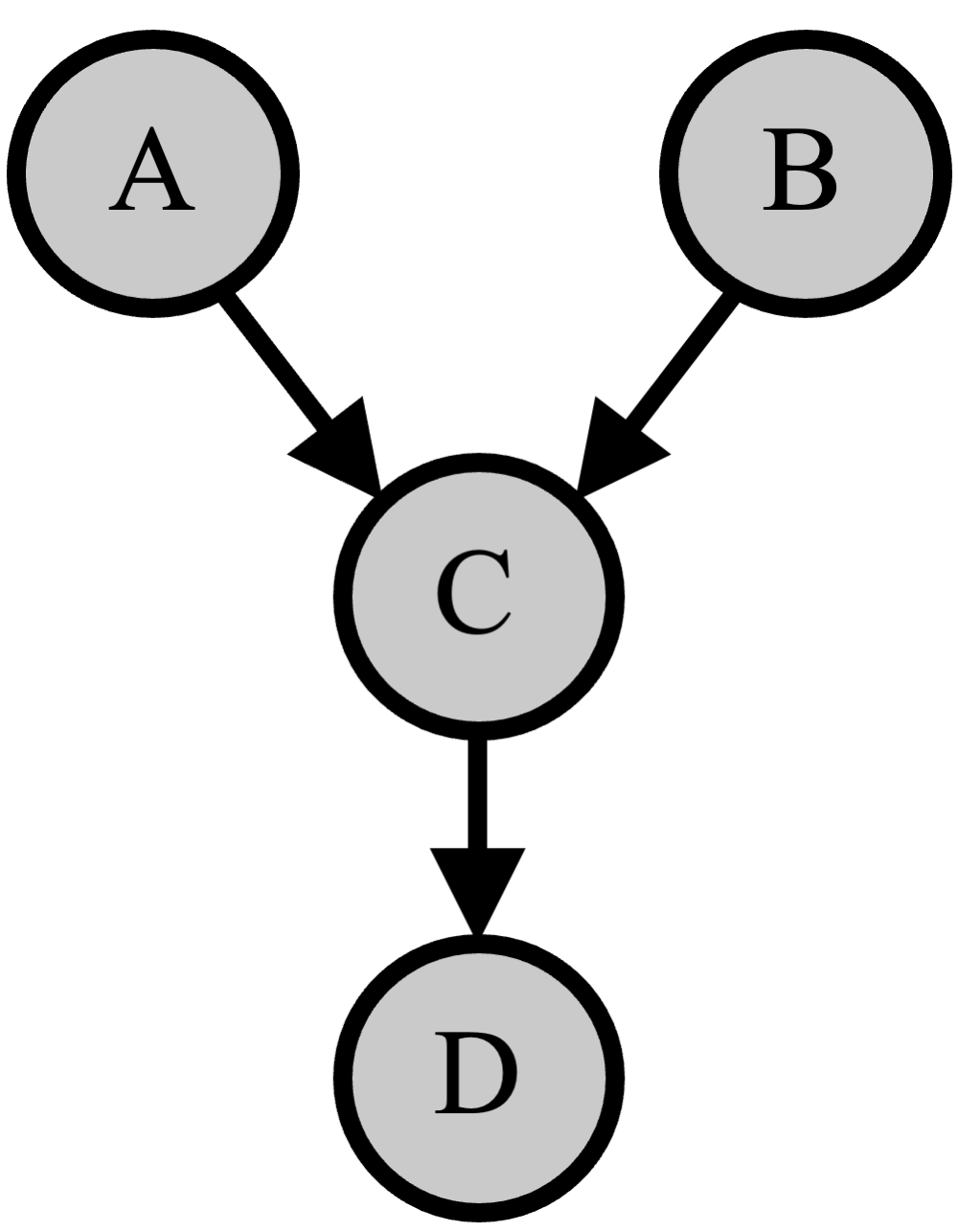}
  \par{(a) Omniscient}
\end{minipage}\hfill
\begin{minipage}[t]{0.16\textwidth}
  \centering
  \includegraphics[width=0.6\linewidth]{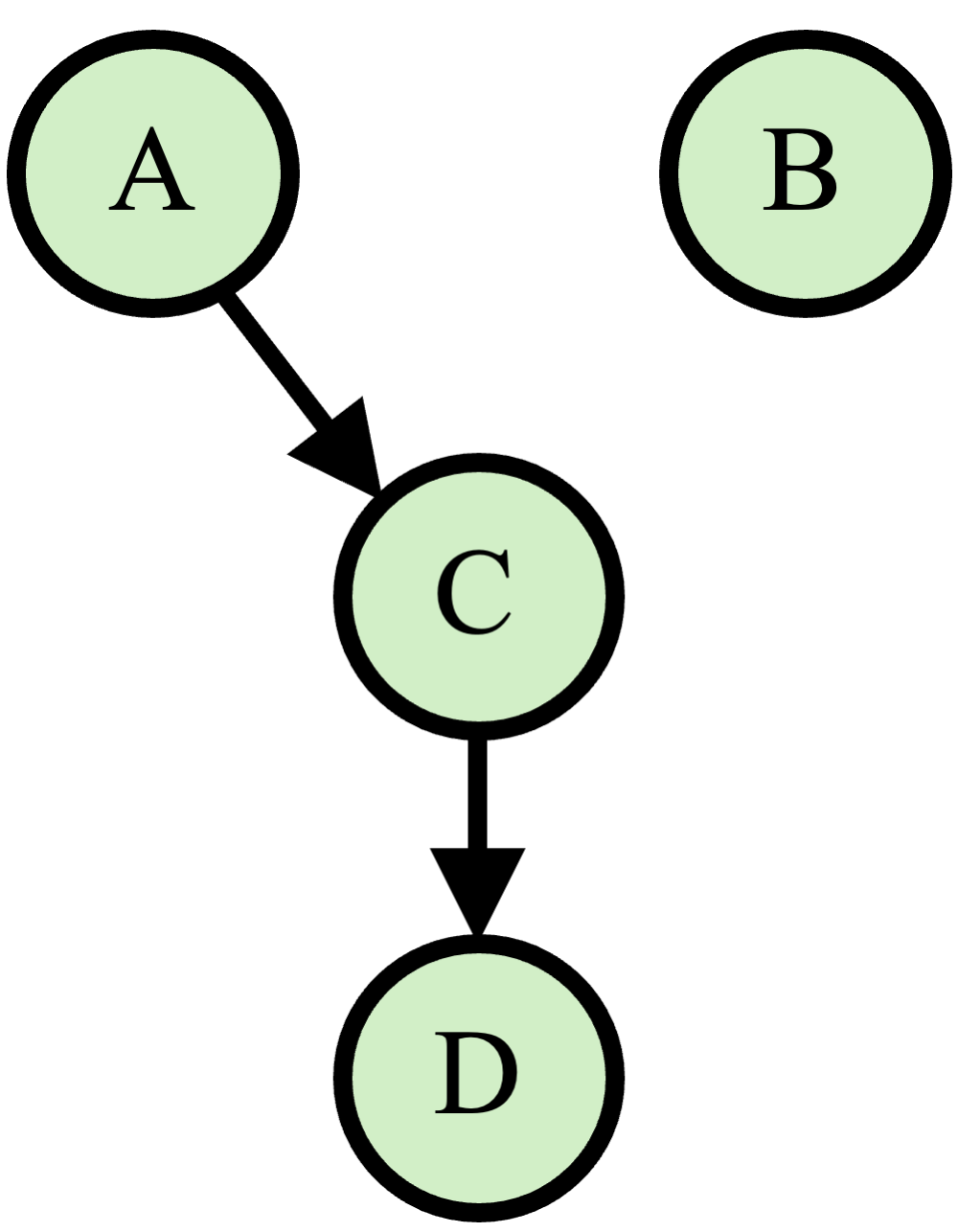}
  \par{(b) Perfect-but-Incomplete}
\end{minipage}\hfill
\begin{minipage}[t]{0.16\textwidth}
  \centering
  \includegraphics[width=0.6\linewidth]{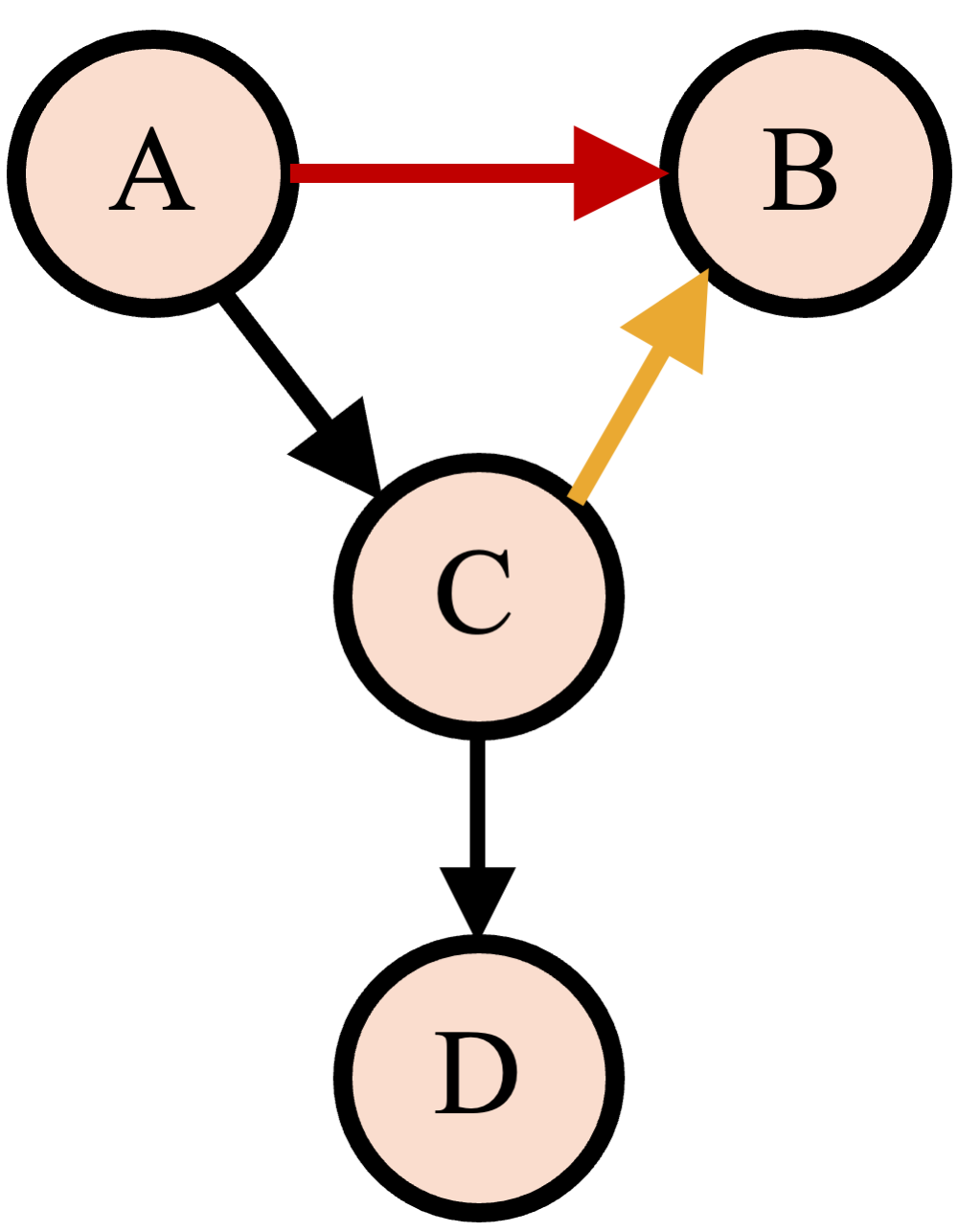}
  \par{(c) Imperfect}
\end{minipage}\hfill
\begin{minipage}[t]{0.16\textwidth}
  \centering
  \includegraphics[width=0.6\linewidth]{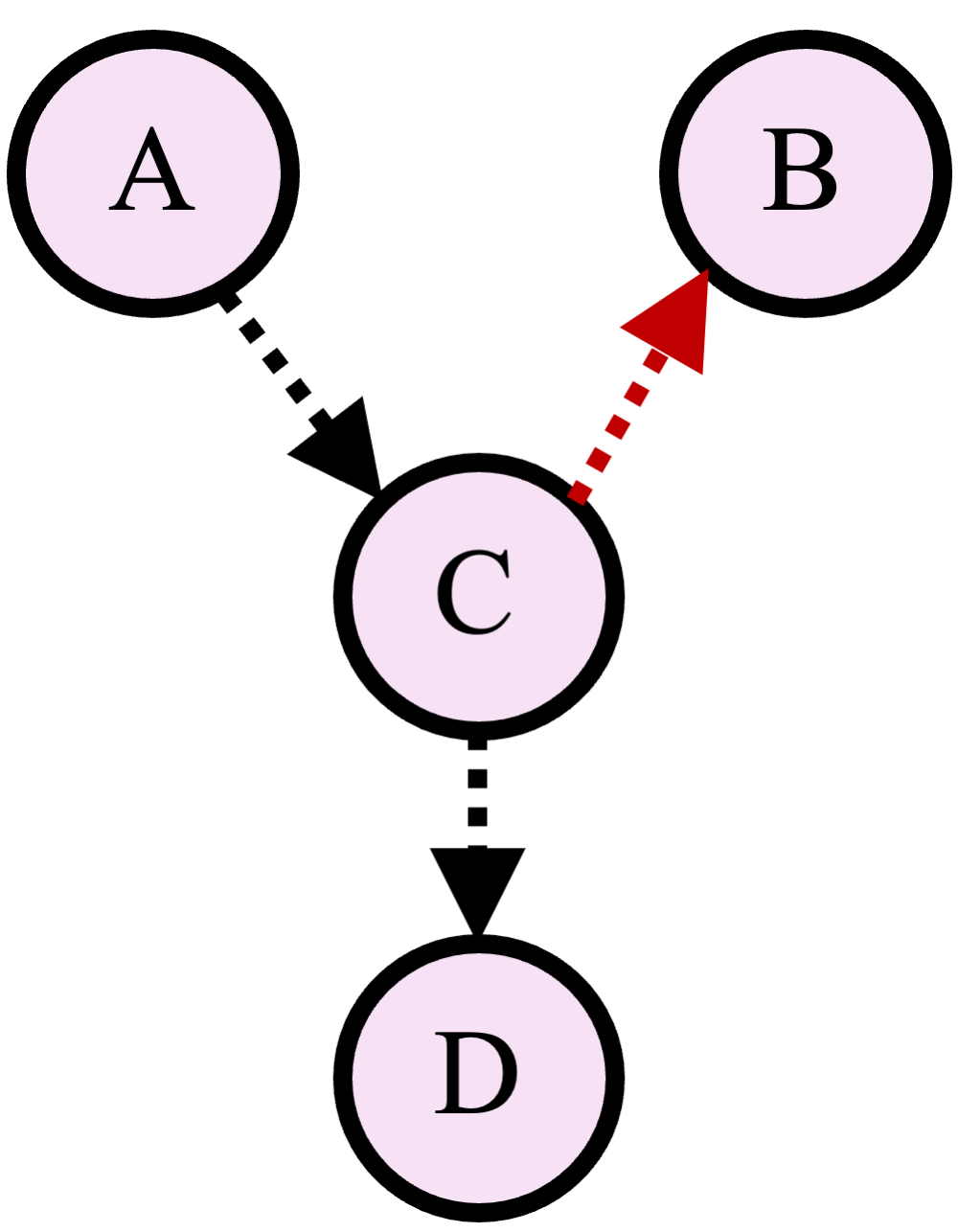}
 \par{(d) Uncertain}
\end{minipage}\hfill
\begin{minipage}[t]{0.16\textwidth}
  \centering
  \includegraphics[width=0.6\linewidth]{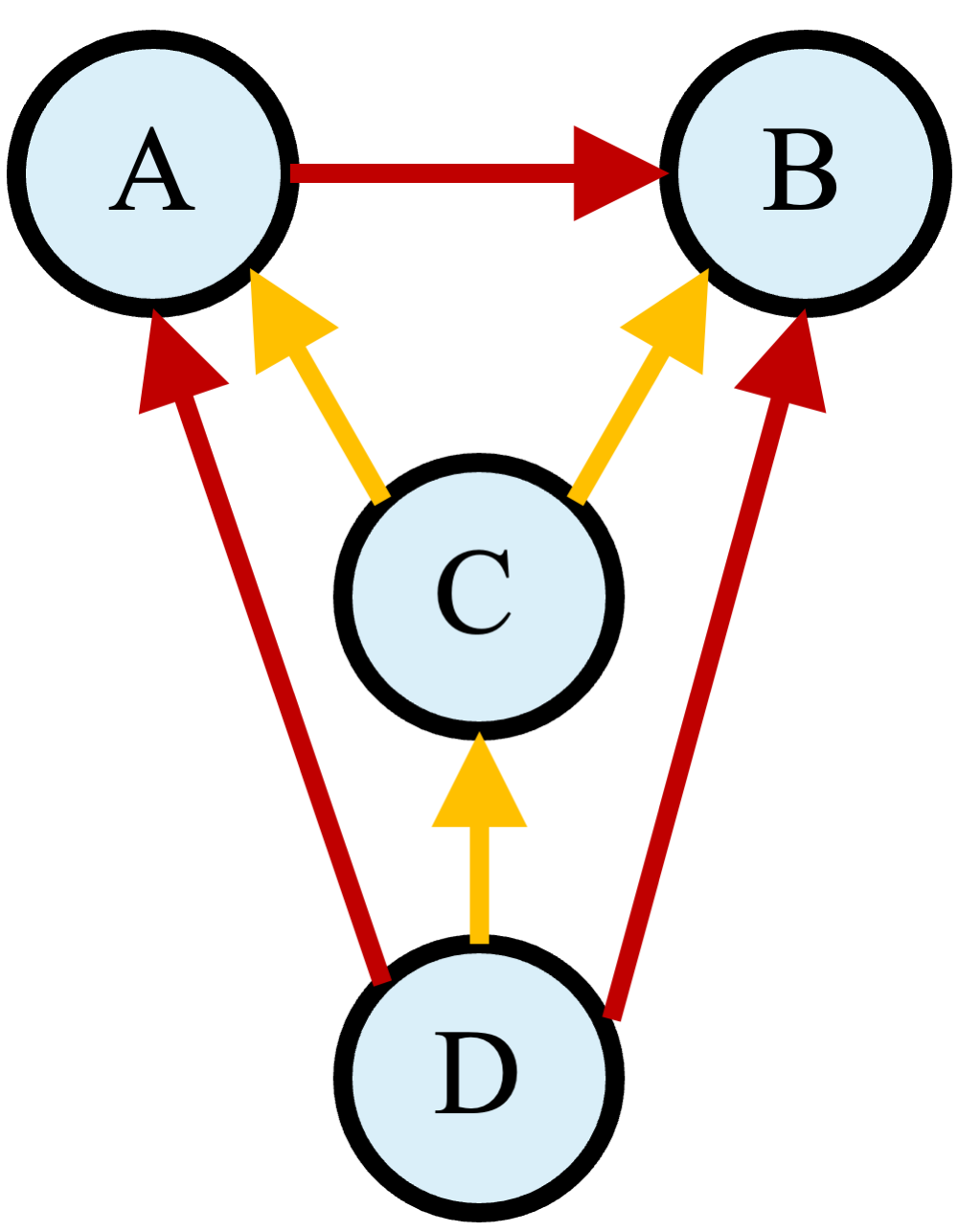}
  \par{(e) Bad Actor}
\end{minipage}
\caption{A toy example illustrating the different expert types. Solid edges represent the causal links asserted by the expert, including correct (black), reverse (yellow) and spurious (red) edges, while dashed edges indicate low-confidence causal links regardless of validity.}
\label{fig:experts_grid}
\end{figure*}

\textbf{Taxonomy of Expert Types.} Based on these dimensions, we provide a few conceptual categories as illustrated in Figure \ref{fig:experts_radar} (see more details in Appendix \ref{sec:radar_demo_details}), and their corresponding behaviors are described through a toy causal graph in Figure \ref{fig:experts_grid}: 
(a) \textit{Omniscient} expert refers to an expert who knows the true causal structure in its entirety with full accuracy. Their knowledge is both complete and perfectly correct as the ground truth. 
(b) \textit{Perfect-but-Incomplete} expert is a weaker type of expert who always provide valid beliefs without asserting a wrong causal link but may not cover all the true links due to gaps in knowledge. 
(c) \textit{Imperfect} experts possess substantial knowledge of the domain but is prone to occasional mistakes (e.g., reversing causal directions or introducing spurious links). Their knowledge is partially correct, typically reflecting a mix of true and false causal beliefs.
(d) \textit{Uncertain} experts are distinguished by their lack of confidence in making judgments on causal relationships. It stands in stark contrast to overconfident experts, who consistently assert strong causal judgments and potentially amplify erroneous links due to unwarranted certainty. 
(e) \textit{Bad Actors} may intentionally inject misleading, adversarial, or strategically manipulated information to distort reasoning or exploit the system, which ultimately undermines the integrity of causal learning. This taxonomy helps clarify assumptions about expert input quality before integrating causal knowledge into models. Notably, a real-world expert often represents a mix of the above types. 

\subsection{Operationalizing the Taxonomy} \label{taxomod}
The value of the proposed taxonomy lies in turning expert heterogeneity from an implicit nuisance into actionable guidance. To enable its use, expert modeling is necessary to infer the latent characteristics that distinguish experts in the crowd. Along the four key dimensions introduced in Section \ref{rwinsights}, these characteristics can be represented by expert-specific parametric models $\boldsymbol{\Theta}$, and learned from patterns in observable individual data. Elicitation therefore serves a dual purpose: collecting causal judgments and generating defining signals about the experts who provide them. While this inference problem is challenging, it connects naturally to several established areas: Completeness measurement requires identifying where an expert has knowledge, which may guide future elicitation effort, and is grounded in knowledge tracing \cite{shen2024survey}, missing data analysis \cite{little2019statistical} and domain adaptation \cite{singhal2023domain}. Belief validity connects to areas such as item response theory \cite{embretson2025item}, truth discovery \cite{li2016survey}, quality control \cite{daniel2018quality} and educational measurement \cite{brennan2023educational}, etc., with the goal of characterizing not only expert accuracy, but also systematic causal error patterns. Another dimension, confidence, provides a natural interface with uncertainty quantification \cite{smith2024uncertainty} and calibration \cite{bozorgzadeh2023model}, allowing for uncertainty-aware quality control over expert judgments that enables structured aggregation. Trustworthiness provides another dimension for modeling expert quality, which further requires tools from robust statistics \cite{loh2025theoretical}, anomaly detection \cite{samariya2023comprehensive}, mechanism design \cite{borgers2015introduction}, and trustworthy AI \cite{li2023trustworthy}. Together, these connections suggest that learning expert types is not a separate preprocessing step, but a core modeling problem that shapes how crowd causal knowledge should be elicited and interpreted.

Once the expert types are determined based on these learned characteristics, they can inform downstream modeling and aggregation by indicating how responses are represented, how much weight they receive, when they should be treated as soft evidence rather than hard constraints, and when additional quality-control or robustness mechanisms are needed. This enables quality-aware use of the wisdom of the crowd, and positions the taxonomy as a bridge between conceptual characterization of causal knowledge and practical mechanisms for elicitation, modeling, and crowd-level aggregation.

\section{Causal Knowledge Models for Eliciting the Crowd}
\label{exp_ml}
As Section \ref{sec:real_exp} illustrates the complexity of crowd causal knowledge, this section is dedicated to modeling and eliciting such knowledge distributed across the crowd. We discuss existing state-of-the-art models and their motivations, applications, and limitations. Take DAG as an example. Let $G = (V, E)$ represent the ground-truth causal structure with $V$ being node set of size $N$ and $E$ being the directed edge set where each $u\to v$ represents a causal link from $X_u$ to $X_v$ for distinct $X_u, X_v \in V$. Consider a crowd of $M$ experts. For each expert $m$, we query $n$ instances, i.e., pairs of variables, to obtain a structured knowledge set $\mathcal{D}_m= \{(\boldsymbol{q}_{i},y_{i})\}_{i=1}^{n}$ where $\boldsymbol{q}_{i} = (u_i, v_i)$ denotes a query over the variables of interest $X_{u_i},X_{v_i} \in V$, and $y_{i}$ denotes the expert response on their causal relationship. The goal is to find the graph that best characterizes the knowledge dataset, the quality of which depends critically on effective knowledge extraction and modeling through elicitation.

\subsection{Expert Causal Knowledge Models for A Single Expert}\label{single_model}
% Let a DAG, $G=(V, E)$, represent the ground-truth causal structure. Here, $V=\{X_1,\cdots,X_N\}$ denotes the set of nodes corresponding to $N$ variables. $E$ is the set of directed edges in $G$ where each $u\to v$ represents a causal link from $X_u$ to $X_v$ for distinct $X_u, X_v \in V$. 
Start from modeling the knowledge of a single expert with $M = 1$. Motivated by cognitive and epistemological studies \cite{swanson1996undiscovered, drayson2022fragmented, katrin2025data}, we assume that the expert has a belief distribution of $G$, denoted as $p(G)$, to describe their understanding of $G$. We use $\tilde{G} = (V, \tilde{E})$ to represent the corresponding graph inferred for the expert with a potential different edge set $\tilde{E}$, i.e., the mean, mode or maximizer of the (posterior) belief distribution $p(G|\mathcal{D})$. For a special case of \textit{omniscient} experts, we have $\tilde{G}=G$. % and eliciting the expert's knowledge is equivalent to revealing the ground truth. 
However, for more realistic scenarios, $\tilde{G}$ deviates from the true $G$ and recovering $G$ solely from the input of a single expert is generally infeasible without additional information. %on their reliability or knowledge level. % Thus, during the elicitation process, our best course is to infer $\tilde{G}$ as accurately as possible without making further assumptions.% \textcolor{red}{Huiling, please check}
To illustrate the analytical formulation, we present two foundational pairwise querying frameworks, differing in the type of causal information extracted and mathematical representation of the queries, namely, the \textit{edge-wise causal knowledge} \cite{amirkhani2016exploiting, bjorkman2025incorporating} and the \textit{ordering-wise causal knowledge} \cite{xiao2018optimal, shojaie2024learning, vashishtha2025causal}.

\textbf{Edge-wise causal knowledge.} This elicitation framework zooms into edges directly and decomposes the expert knowledge graph $\tilde{G}$ into a set of local edge-level beliefs. % Each query $\boldsymbol{q}_i$ includes a pair of different nodes $X_{u_i}, X_{v_i}\in V$ where $1\leq u_i<v_i\leq N$, and 
For each query $\mathbf{q}_i$, the expert provides the indicator $y_i \in\{1,0,-1\}$ as their belief in the directed causal effect between the corresponding variables $X_{u_i}$ and $X_{v_i}$, corresponding to three cases: $u_i\to v_i$, no link, $v_i\to u_i$. It follows a categorical distribution over three outcomes with parameters $\boldsymbol{\theta}_{u_i, v_i}$ conditioned on $\boldsymbol{q}_i$. Given such a knowledge set $\mathcal{D}$, eliciting a causal graph from expert can be formulated as $p(G, \boldsymbol{\Theta}|\mathcal{D}) \propto % p(\mathcal{D}|\tilde{G}, \boldsymbol{\Theta}) p(\boldsymbol{\Theta}|\tilde{G})p(\tilde{G}) \\
\prod_{i = 1}^n p(y_i|\mathbf{q}_i, \boldsymbol{\Theta})p(\boldsymbol{\Theta}|G)p(G)$ where $\boldsymbol{\Theta}=\{\boldsymbol{\theta}_{u,v}\}_{1\leq u<v\leq N}$ is a set of learnable parameters that characterize the edge beliefs of expert,  % It should encode the expert's characteristics, 
such as domain relevance, confidence, and behavioral uncertainty discussed in Section \ref{rwinsights}. % Examples of the information it may encode includes edge-specific domain relevance, confidence, and behavioral uncertainty. % (a) edge-specific domain relevance, which captures how well a particular edge falls within the expert's domain expertise and can be modeled as a Bernoulli parameter to indicate the likelihood of a correct belief in the edge, (b) general behavioral uncertainty, which can be reflected by a variance parameter shared by all the edges, and (c) confidence in their belief in the edge, which can be represented by edge-specific factors that modulate the general uncertainty. 
% Rather than committing to a fixed formulation, we keep the flexibility to accommodate various expert types depending on the contexts. 

% \begin{align*}
% \max_{\tilde{G}\in\mathcal{G}}p(\tilde{G}|\boldsymbol{\theta})
% \;\; s.t. \;\;\boldsymbol{\theta}\in\mathop{\arg\max}_{\boldsymbol{\theta}\in\mathbf{\Theta}}\,p(\boldsymbol{\theta})\prod_{i=1}^np(y_i|\boldsymbol{\theta},\boldsymbol{q}_i)\quad 
% \end{align*}
% \textcolor{red}{$y_i$ socre and }
% where $\boldsymbol{\theta}\in\mathbf{\Theta}$ denotes the model parameters associated with the edges to learn and $p(\boldsymbol{\theta})$ is the corresponding prior over the parameter space $\mathbf{\Theta}$. 

\textbf{Ordering-wise causal knowledge.} The knowledge graph $G$ essentially induces a topological ordering over the corresponding variables. 
% It inspires us to learn the expert's beliefs and infer the global structure of $G$ from ordering information rather than direct edge information, thereby enabling broader and more flexible structural constraints with less computational complexity and more data efficiency. 
For the expert knowledge graph $\tilde{G}$, its causal order function $\pi$ satisfies that, for any variable pair $X_u,X_v\in V$ with a directed path from $X_u$ to $X_v$, $\pi(X_u)<\pi(X_v)$, i.e., smaller values indicate higher ranking or being more upstream in the causal flow.  %With this language, the causal flow always propagates from a variable with a higher ranking to another variable with a lower ranking. 
To elicit experts with ordering information in this framework, $y_i$ provides a scaled score to convey the expert's belief in their ordering relationship instead. %, namely, which variable is more upstream and by what strength. To learn the causal knowledge graph with the query data $\mathcal{D}$, unlike the edge-wise setting, it is difficult to admit the same parametrization $\boldsymbol{\theta}$ due to the complexities of the ordering relationships. 
Inspired by \cite{burges2005learning, cao2007learning}, we assume a latent score function $\phi(\cdot)$ as a learnable surrogate of the order and $\Phi=[\phi(X_1),\cdots, \phi(X_N)]$, assigning each variable a utility score to determine their order in a path, % the higher value of which indicates a higher ranking, namely
satisfying $\pi(X_u)<\pi(X_v)$ implies $\phi(X_u)>\phi(X_v)$. % By introducing an incidence-like matrix $\mathbf{B}=\left(b_{i,j}\right)_{n\times N}$ where $b_{i,u_i}=1$ and $b_{i,v_i}=-1$ for $\forall i=1,\cdots,n$, and other elements all being zero, 
The posterior can be obtained similarly as 
% \begin{align*}
% \min_{\Phi}||\boldsymbol{y}-\mathbf{B}\Phi||_2^2\;\; s.t.\;\; g(\Phi)\leq 0,\; h(\Phi)=0
% \end{align*}
% \begin{equation*}
$p(G,\Phi,\boldsymbol{\Theta}|\mathcal{D})\propto  \prod_{i=1}^np(y_i|\boldsymbol{q}_i,  \Phi,\boldsymbol{\Theta})p(\Phi|G,\boldsymbol{\Theta})p(\boldsymbol{\Theta}|G)p(G) $
% &\min_{\Phi,\sigma^2}p(\Phi|\sigma^2, \boldsymbol{y},\mathbf{B})p(\sigma^2|\boldsymbol{y},\mathbf{B})\;\; s.t.\;\; \boldsymbol{y}\sim \mathcal{N}(\mathbf{B}\Phi,\sigma^2\mathbf{W^{-1}})
% \end{equation*}
% where $\boldsymbol{y}=[y_1,\cdots,y_n]$,  $\Phi=[\phi(X_1),\cdots, \phi(X_N)]$, and 
with $\boldsymbol{\Theta} = \{\boldsymbol{\theta}_{u,v}\}_{1\leq u<v\leq N}$ denoting latent parameters characterizing the expert's beliefs. % in the pairwise relationships (upstream, downstream or no path) between variables. % . Similar to the edge-based case, 
% We use $\boldsymbol{\Theta}$ to encode the expert's characteristics including domain relevance, global uncertainty, pairwise confidence, etc.

\textbf{Comparison for different experts: Edge v.s. Ordering. }
Edge-wise causal knowledge captures the most fine-grained local structure, while ordering-wise causal knowledge emphasizes more on the global structure of $G$. For an \textit{omniscient} or \textit{perfect-but-incomplete} expert, edge-wise elicitation directly reveals a portion of the ground truth and theoretically guarantees the full recovery through exhaustive queries for \textit{omniscient} experts. However, the explicitness limits the inferential flexibility as edges provide only specified information, whereas
ordering-wise knowledge enables broader structural inference. 
Such granularity renders the elicitation of edge-wise knowledge inherently demanding, whereas ordering-wise knowledge is often more query-efficient for the inference of $G$. 
With imperfect experts, edge-wise framework struggles with incomplete or biased knowledge or even introducing acyclicity violation due to their incompleteness, uncertainty, and thus becomes a risky choice. However, more robustness can be gained from ordering-wise knowledge in this case as it accommodates graded confidence through its scoring mechanism, enabling systematic detection of belief patterns and providing more diagnostic information about inconsistency or uncertainty.
These frameworks present a fundamental trade-off between knowledge efficiency and structural specificity. 

% \vspace{-1.5em}

\subsection{Leveraging the Wisdom of the Crowd} \label{crowd}
Now we extend from a single expert to a crowd and highlight the importance of leveraging collective intelligence as a powerful driver for causal learning with potential approaches to enable this at scale. 

% \subsection{Harnessing the Crowd Knowledge}
Quantitative expert-driven paradigms are promising to handle the complexities in causal discovery. However, reliance on a single expert is inherently fragile, as individual judgments are inevitably partial, biased, and susceptible to mistakes. The true potential of expert knowledge emerges in its collective form where diversity across a crowd of experts yields both mutual reinforcement through agreement and complementary perspectives through disagreement. This combination enables crowd knowledge to be more robust to errors and encompass a spectrum of insights. Therefore, a shift from expert-in-the-loop to crowd-in-the-loop causal learning is warranted, moving beyond the authority of one toward the collective wisdom of many.
% Suppose we have a crowd of $M$ experts with knowledge datasets $\mathcal{D}_{1},\cdots,\mathcal{D}_M$. 

Generally speaking, there are at least two strategies with different modeling philosophies for leveraging collective knowledge: (a) \textit{expert-level aggregation}, which emphasizes the refinement of individual beliefs $\boldsymbol{\Theta}_1,\cdots, \boldsymbol{\Theta}_M$ and their aggregation; and (b) \textit{query-level aggregation}, which learns a collection of query-level models with hyperparameter $\boldsymbol{\Theta}_{u,v|G}^{m, k}$ for pair $(u,v)$ of expert $m$ in scoring class $k$. %, and then characterizes the beliefs with $p(G|\boldsymbol{\Theta})$. 

% \ryan{the use relies on: 1) denoise and generalize - refine individual data with crowd information 2) integrate: either first fuse data then learn global model, or first individual models then fuse.  same query or not? 3) bad actor detection}\textcolor{red}{I'll fix this shortly} % \textcolor{purple}{distributional assumption/estimation for $P(K)$, network structure?}

\textbf{Expert-Level Aggregation }is conceptually straightforward, the final goal of which centers on two questions \textit{how to build high-quality individual models} and \textit{how to aggregate the models to generate crowd-based beliefs}. $\cup_{m=1}^M\boldsymbol{\Theta}_m$ can be combined through methods ranging from simple average to sophisticated probabilistic approaches that account for the distribution of expert characteristics. With the human causal knowledge models described in Section \ref{single_model}, one example can be formulated as follows: $p(G, \boldsymbol{\Theta}| \cup_{m = 1}^M\boldsymbol{\Theta}_m) \propto \prod_{m=1}^M p(\boldsymbol{\Theta}_m | G, \boldsymbol{\Theta})p(G, \boldsymbol{\Theta}). $
While intuitive, this approach requires committing to individual graph structures before aggregation, the performance of which critically depends on the individual causal graph inference and aggregation approach, potentially discarding valuable information about expert uncertainty.  

\textbf{Query-Level Aggregation } % This approach models the response generation process directly without requiring inference for individual causal graph, %query-level fidelity through crowd-based aggregation, which
% offering greater flexibility for crowd-based aggregation. %alternative by modeling the response generation process directly without inference of individual causal graphs. 
% provides a principled path from individual expert responses to collective inference about causal structure, explicitly modeling both the cognitive limitations of experts and varying difficulty of queries while remaining agnostic to the specific elicitation format employed. 
recognizes that expert responses emerge from a complex mixture of cognitive processes rather than direct observations of truth. For example, 
%We propose a probabilistic framework that infers $G$ from $P(y_{m,i} |G)$ of all queries $i \in \{1, \ldots, n\}$ and experts $ m \in \{1,\ldots, M\}$.  
for any pair of variables $(u_i, v_i)$, we can model the score of expert $m$, $y_{m,i}$, as arising from one of three latent mechanisms corresponding to the upstream evidence (positive score), downstream evidence (negative score), or no evidence (weak score). % A latent indicator $z_{u_i,v_i}^{(m)} \in \{+, - ,0\}$ indicates which mechanism generates the observed score, with mixing probability $p\left(z_{u_i,v_i}^{(m)} = k\right) = \pi_k$ satisfying $\sum_{k\in\{+, -, 0\}}\pi_k = 1$. % This representation aligns naturally with expert reasoning about causal mechanisms. 
Conditional on this mechanism, score can be characterized as % $y_{m,i} |G, z_{u_i,v_i}^{(m)} = k$ which follows the distribution $p_{k}\left(y;\boldsymbol{\theta}_{u_i,v_i|G}^{(m,k)}\right)$ where $\boldsymbol{\theta}_{u_i,v_i|G}^{(m,k)}$ is the hyperparameter for $(u, v)$ in class $k$ of the $m$th expert. The concrete form of 
a realization from a mixture distribution, the concrete form of which depends on whether specific edge-wise or ordering-wise knowledge elicitation employed. % Under edge-wise framework, experts make discrete judgments about whether a directed edge exists or not $y_{m,i}\in \{0,1 \}$, while under the ordering-wise framework, experts provide quantified judgments about causal precedence rather than discrete edge existence through a score $y_{m,i} = \phi_m(u_i, v_i)$. 
Some mechanism-specific decomposition can be applied to $\boldsymbol{\theta}_{u_i,v_i|G}^{(m,k)}$ such as $f_{G}^{(m)}(u,v)\times g(u,v|G)$, which separates two fundamental variation sources: a query-specific component $g(u,v|G)$ captures the intrinsic difficulty of identifying the relationship between $X_u$ and $X_v$ in $G$, independent of expert identity while a expert-specific component $f_{G}^{(m)}(u,v)$ will further modulates the distribution depending on the characteristics of the expert. %response based on their %knowledge level, domain familiarity, and behavioral 

\textbf{Discussion.} These two strategies entail distinct trade-offs in computational efficiency, modeling flexibility, and information use. Expert-level aggregation is a modular two-stage process with computational advantages, yet inherits all challenges of individual-level causal graph inference. Additionally, collapsing diverse structures into a single graph may obscure valuable information. While expert characteristics can be incorporated through weighted average or distributional assumptions, specifying appropriate forms for optimally combining graphs of varying quality remains challenging. Query-level aggregation bypasses individual-level inference by directly modeling responses toward the global $G$, preserving richer information and naturally accommodates cognitive mechanisms by explicitly decomposing query difficulty with higher computational cost due to a larger parameter space. Both frameworks benefit from higher proportions of knowledgeable and reliable experts, but may handle heterogeneity distinctly: expert-level aggregation struggles with contradictory individuals, while the query-level better leverages fragmented knowledge through fine-grained response modeling. 
% \vspace{-1.5em}
\subsection{Extension of Elicitation Frameworks}

Beyond pairwise elicitation, these frameworks extend to richer forms of causal knowledge. Graph-wise elicitation elevates the query unit from individual edges to entire graphs or subgraphs, enabling experts to propose candidate structures or coherent structural constraints. List-wise elicitation instead captures beliefs about causal flow through ranked variable orderings, such as triplets or quadruplets \cite{vashishtha2025causal}. These higher-order schemes provide experts greater expressive flexibility, but they also differ in informational content and elicitation difficulty. A complementary distinction lies between insertion- and deletion-based elicitation. Deletion-based schemes are theoretically attractive because, under the Markov condition, removed edges directly encode conditional independencies through $d$-separation. In contrast, inserted edges require additional faithfulness, acyclicity, and consistency checks to distinguish asserted relations from unexamined pairs. In practice, however, insertion-based elicitation is often simpler and more scalable, particularly for sparse graphs, since experts typically find it easier to enumerate known relations than exhaustively eliminate implausible ones. Balancing theoretical informativeness with practical elicitation efficiency therefore remains an open challenge.

\section{Crowd Quality Control}\label{sec:QC}
\subsection{Quality Control against Spurious Correlations}
Leveraging the wisdom of the crowd does not imply relying on an unrestricted general public. In many realistic settings, there are structured inclusion criteria used to recruit professionals, trainees, or domain-informed contributors. Our framework is compatible with such structured crowds and does not assume purely lay participation. However, distinguishing true causal relationships from widely held spurious beliefs is still challenging. Spurious beliefs often appear as detectable response patterns, or more specifically, internal inconsistencies across queries on related variable pairs. Thus, crowd-based causal learning may recognize such unreliable patterns through taxonomy-based expert modeling and appropriate query design (e.g., interventional question framing, repeating queries, or counterfactual probes), and down-weight them with quality-aware aggregation. When crowd signals are weak or reflect shared misconceptions, external sources such as LLMs can provide complementary sanity check. More broadly, these quality control mechanisms are far from mature, which offers an open opportunity for crowd-based causal learning.

\subsection{Optimal Design of Elicitation}
\label{bn_doe}
%- formulation: optimal design + agent based
In practice, expert elicitation is constrained by time, budget, labor, and cognitive load, making exhaustive querying of all variable pairs infeasible. This motivates viewing elicitation through the lens of experimental design and active learning, where the goal is not to ask more questions, but to ask the most informative ones. Given an inferred graph $\tilde{G}$ and candidate query pool $\mathcal{C}=\{(u,v)\}_{1\leq u<v\leq N}$, the elicitation process can be formulated as a sequential optimal design problem that adaptively selects variable pairs to maximize information gained under limited interaction budgets. At each stage $t$, a design $\xi_t$ allocates a limited budget $K_t$ across candidate queries according to an information utility $U(\xi_t)$, which may be defined through classical criteria such as E-optimality \cite{xiao2018optimal}, expected information gain (EIG) \cite{bjorkman2025incorporating}, or other uncertainty-reduction objectives. For reliable experts, queried pairs may be removed as knowledge accumulates; for noisy or uncertain experts, repeated querying may instead be beneficial. The key point is not a specific optimization criterion, but the broader possibility that expert elicitation can be transformed from a heuristic process into a principled quality-aware design framework that strategically reduces uncertainty about the causal graph under realistic resource constraints. This perspective opens a rich set of research questions at the intersection of causal discovery, active learning, uncertainty quantification, and quality management and control.

\subsection{Agent-based Expert Simulation}
Scaling causal learning to large crowds faces practical constraints including labor cost, fatigue, and unreliable participants, motivating the use of surrogate agents that emulate expert reasoning while remaining scalable and controllable. Recent advances in LLMs provide a promising foundation, as they encode broad commonsense knowledge and can generate plausible causal structures with minimal querying \cite{shaposhnyk2025can,long2023causal,kiciman2023causal}. Beyond language, foundation models, world models, and multi-agent systems further enable simulation of collective reasoning and large-scale digital populations \cite{jin2024genegpt,audenaert2025causal,richens2024robust,ding2025understanding,li2024simulating,zhangkabb2025kabb,lin2025crowdllm,berenberg2018efficient}. This motivates a hybrid crowd framework in which human and simulated agents jointly contribute to causal learning. However, LLM agents should be treated as imperfect experts due to hallucination, instability, and associative bias; thus safeguards such as consistency checks, cross-model agreement, benchmark calibration, and human-in-the-loop verification remain essential, with LLM outputs aggregated as complementary rather than dominant signals.

\section{Alternative Views} \label{altv}

A natural objection is that \textit{expert knowledge should remain only a supplement to observational or interventional data rather than a central component of causal discovery}. We do not argue that expert knowledge replaces data or is inherently superior. Instead, we view it as an increasingly important complementary source of information. Observational and interventional approaches often face limitations in identifiability, cost, computation, or ethics \cite{eberhardt2017introduction}. In contrast, well-elicited expert knowledge can introduce domain inductive biases unavailable from data alone \cite{hasan2026dkc, zhou2013incorporating}, reduce hypothesis spaces, and provide interpretable guidance for causal learning. Under severe data scarcity, structured expert knowledge may even support causal discovery when observational evidence is limited or unavailable. Another concern is that \textit{causal learning should rely only on carefully vetted experts rather than broader crowds}. While expertise quality is crucial, restricting participation to a small set of elite experts may reduce diversity and scalability \cite{yoo2024elicitation, hasan2024boosting}. We instead advocate inclusive elicitation combined with quality control, denoising, and model-based aggregation, shifting reliability from expert pre-selection to principled post-hoc integration.

We also acknowledge important practical and ethical challenges, including recruitment, incentives, adversarial behavior, informed consent, privacy, misinformation, and bias in knowledge representation \cite{schlagwein2019ethical, standing2018ethical,zdravkova2020ethical}. Fortunately, many of these issues have been extensively studied in crowdsourcing, HCI, and responsible AI. Techniques such as rigorous recruitment, fair compensation, reputation systems, incentive design, and anomaly detection help mitigate low-quality or adversarial contributions while preserving diversity. Crowd-based causal learning should therefore be developed not merely as a technical framework, but as a socio-technical system requiring interdisciplinary solutions.

\section{Conclusion} \label{act}
In this paper, we highlight the opportunities for crowd-based causal learning as a promising paradigm to leverage distributed human expertise in causal discovery. We call on the community to advance this paradigm by developing new methodologies, benchmarks, tools, and theoretical foundations. Table~\ref{tab:future_directions} in Appendix~\ref{future} highlights several concrete avenues for future exploration. These include modeling the nuanced knowledge profiles and uncertainty patterns of human contributors; optimizing elicitation strategies to make crowd input more reliable, cost-effective, and scalable; and designing hybrid workflows where human and AI systems, especially LLMs, collaborate seamlessly. Beyond technical methods, this agenda raises foundational theoretical questions about the limits and guarantees of distributed causal learning, along with pressing system-level challenges in evaluation, ethics, and inclusivity. Fully realizing the impact of the paradigm will require deep interdisciplinary collaboration that draws on insights from causal inference, HCI, crowdsourcing, AI ethics, and cognitive science to build robust, adaptive, and human-centered causal learning systems.

% \textbf{Conclusion.} In this paper, we take the position that causal learning should embrace the wisdom of the crowd in recognition of the rich and diverse knowledge accumulated by humans over time. We outline a unified framework to realize this vision through knowledge elicitation, principled process optimization, and crowd-based integration and decision-making, supported by concrete examples and experiments. We reimagine causal learning not as a purely data-driven task but as a scalable and distributed human-in-the-loop process that elevates diverse human expertise as a central pillar of causal learning pipelines. We hope this new paradigm will resonate beyond across the broader scientific community and emerge as a compelling direction that catalyzes the next generation of causal learning in the era of AI.

\bibliographystyle{unsrt}
\bibliography{example}

\newpage
\onecolumn

\begin{appendices}
\section*{\Large Appendix}
\vspace{0.5cm}

\section{Glossary}
\begin{definition}[Path]
A path in a directed graph $G = (V, E)$ is a sequence of nodes $v_1, v_2, \ldots, v_k$ with the property that each consecutive pair $v_i$, $v_{i+1}$ is joined by a
directed edge $(v_i, v_{i + 1})\in E$. 
\end{definition}
\begin{definition}[Directed Acyclic Graph (DAG)]
A directed acyclic graph (DAG) $G$ is a directed graph with no directed cycles. In other words, there is no path that starts and ends at the same node.
\end{definition}

\begin{definition}[Parent variable]
A parent variable $X_u$ of an variable $X_v$ in a directed graph $G$ is its direct cause that has an directed edge linking from $X_u$ to $X_v$.
\end{definition}
\begin{definition}[Bayesian Network]
A Bayesian network is the representation of a probability distribution on a DAG $G=(V,E)$ . It specifies a joint distribution over $V=(X_1,\cdots, _N)$ as a product of local conditional distributions, i.e., $P(X_1,\cdots, X_N)=\prod_{i=1}^Np(x_i|\text{Pa}_G(x_i)).$
\end{definition}

\begin{definition}[Topological order]
A topological order of a directed graph $G = (V, E)$ is an ordering of its nodes as $v_1, v_2, \ldots, v_k$ so that for every edge $(v_i, v_j)$ we have $i < j$.
\end{definition}

\section{Background}

\textbf{History of Causal Knowledge Elicitation.} As summarized in Table \ref{tab:expert_roles}, the incorporation of human expertise into causal models is not a new endeavor, but has historically been marked by heuristic, unscalable practices \cite{o2019expert}. In the early days of BN research, the construction of the DAG structure was largely a manual process \cite{heckerman1992toward, jackson1990introduction}. Experts would work with knowledge engineers to define variables of interest, identify their states, and draw directed edges based on professional experience. A well-known example is the medical belief network \textit{ALARM} \cite{beinlich1989alarm}. As the fields of statistics and ML came of age, the focus shifted toward data-driven structure learning. Algorithms were developed to find the structure that maximized the marginal likelihood of data or satisfied conditional independence constraints \cite{ramirez2001building, neapolitan2004learning}. However, purely data-driven methods cannot distinguish between structures in the same MEC with observational data. To resolve such ambiguities, researchers turned to intervention-based strategies that often require strict experimental conditions and employ costly operations \cite{Frederick2008intervention, li2023causal,choo2024causal}. They can also integrate expert knowledge as ``side information'' for disentanglement \cite{amirkhani2016exploiting, gonzales2022hybrid}. Yet hybrid approaches treat expert knowledge in a largely centralized manner through consulting single experts or small specialist groups whose models inherit individual biases and lack collective intelligence. Moreover, these methods often overlook the fact that experts might only have local knowledge of specific sub-domains. The absence of systematic framework for decentralized intelligence solicitation and consolidation has hindered scaling causal discovery to meet modern big data demands. 

\section{Details of the Radar Chart in Figure 4}
\label{sec:radar_demo_details}

To illustrate the difference among expert knowledge and behavior discussed in the main text, we simulate a radar chart that visualizes different expert types along four dimensions: \emph{Completeness}, \emph{Belief Validity}, \emph{Confidence Level}, and \emph{Trustworthiness}. We consider five representative expert types to illustrate distinct patterns of knowledge quality and behavior. (I) The \emph{omniscient expert} is an idealized upper bound that achieves near-perfect scores along all four dimensions. It serves purely as a conceptual reference rather than a realistic human expert. (II) The \emph{perfect-but-incomplete expert} exhibits high belief validity, confidence level, and trustworthiness, but low completeness, which means this type of experts is highly reliable within his domain knowledge but lacks broad knowledge reserves. (III) The \emph{imperfect expert} represents individuals with relatively broad coverage but only moderate belief validity. (IV) The \emph{uncertain expert} exhibits high validity and trustworthiness but low expressed confidence, reflecting conservative behavior under uncertainty. (V) The \emph{bad actor} corresponds to adversarial or uncooperative behavior, characterized by low belief validity and low trustworthiness despite potentially high completeness or confidence.

% \textcolor{brown}{\textbf{(Can add maths notations?)} To compare the performance of different types of human experts, we characterize human experts by the amount and quality of structural information they provide, regardless of the specific learning algorithm. Specifically, an expert's knowledge can be analyzed along four statistical dimensions: (I) Specificity (1-FPR), measured by the presence of incorrect edges regardless of direction; (II) Consistency (Acyclicity), reflected by the number of directed cycles induced by the provided edges, which determines how many edges must be removed to obtain an acyclic graph; (III) Sensitivity(Coverage), defined as the fraction of true edges that are mentioned, ignoring direction; (IV) Directional accuracy \underline{(1-Reverseness)? did not define reverseness}, quantified as one minus the edge reverseness rate; and (V) Uncertainty awareness: experts' different levels of confidence about their scores.}

\section{Details of the Real-World Case Study}\label{survey}
This section presents more details of the real-world case study in Section \ref{real_exp}, including ethical approval and the survey design used to collect participant responses.

\subsection{Study Ethics and IRB Approval}\label{irbs}

In our study, although the participants were only asked to provide causal annotations without sensitive information (e.g., demographics), their responses may be still informative about the participants themselves (e.g., their knowledge, expertise or behaviors). Inspired by the ongoing discussions on human-subjects status and ethics of crowdsourced annotation \cite{shmueli2021beyond, kaushik2024resolving, xia2022original}, we treat this study as human-subjects research rather than purely mechanical annotation work, and follow the corresponding IRB procedure at our institution as part of responsible research practice. Our study protocol was reviewed by the IRB, and was determined to be exempt
status (Category 2 - Educational tests, surveys, interviews, observations of public behavior). The study is anonymous and involves no more than minimal risk to the research subjects. Participants were recruited and completed the survey online. They were informed of the study purpose, the voluntary nature of participation, and how their responses would be used. The study collected only pairwise causal direction and strength judgments among variables in a benchmark Bayesian network \textit{Asia}, and no sensitive or personally identifiable information was collected. All responses were analyzed for research purposes only; both aggregate-level and participant-level results were reported, and any participant-level visualizations were anonymized.

\subsection{Survey Design}\label{survey}
Now we demonstrate the details of the real world questionnaire used for causal elicitation in our case study, where we consider two elicitation protocols: \emph{edge-based} and \emph{ordering-based}.

In \textbf{edge-based elicitation}, participants were presented with the variable pair $(A,B)$ and asked to indicate whether there is a \emph{direct} influence between them. Responses were recorded as a ternary integer score $r_{AB} \in \{-1,0,1\}$, where $1$ denotes $A \rightarrow B$, $-1$ denotes $A \leftarrow B$, and $0$ indicates no perceived direct causal influence. 

In contrast, the \textbf{Ordering-based elicitation} asks participants to assess the \emph{total causal influence} of $A$ on $B$, including both direct and indirect effects.
Participants rated their belief on an integer scale from $[-10,10]$, where positive ratings indicate participants' belief in $A \rightarrow B$, negative indicate the opposite situation, and zero indicates no perceived causal relationship.

To make the distinction concrete, consider the \textit{Asia} network as an example. In this network, \textbf{Smoking} has a direct causal effect on \textbf{Lung Cancer}, so an edge-based question for the pair $(\textbf{Smoking}, \textbf{Lung Cancer})$ asks whether participants believe there is a direct edge between the two variables. A response of $1$ indicates the perceived direct relation $\text{Smoking} \rightarrow \text{Lung Cancer}$, while a response of $0$ indicates that no direct causal influence is perceived.

By contrast, ordering-based elicitation asks whether one variable is causally upstream of another, regardless of whether the effect is direct or mediated by intermediate variables. For example, \textbf{Smoking} may be viewed as upstream of \textbf{Dyspnea} through causal paths involving lung disease variables. For the pair $(\text{Smoking}, \text{Dyspnea})$, a participant may assign a positive score even if they do not believe there is a direct edge from \textbf{Smoking} to \textbf{Dyspnea}. The magnitude of the score reflects the strength of this perceived upstream causal influence.

The survey included examples showing how causal beliefs should be mapped to signed numerical ratings. For instance, one ordering-based question asked:
\begin{quote}
\emph{How strongly do you believe that \textbf{Smoking} is an upstream causal variable of
\textbf{Lung Cancer}?}
\end{quote}
Each participant provided a single integer-valued response. Across the population,
the collected responses form a set of signed scores
$\{ r^{(m)}_{\text{Smoking},\text{Cancer}} \}_{m=1}^M$.
A comparison between the two elicitation protocols is summarized in
Table~\ref{tab:survey_comparison}.

\begin{table}[!h]
\centering
\caption{Comparison between \textbf{edge-based} and \textbf{ordering-based} elicitation.}
\label{tab:survey_comparison}
\vspace{1em}
\resizebox{\linewidth}{!}{
\begin{tabular}{|p{3.2cm}|p{5.2cm}|p{5.2cm}|}
\hline
\textbf{Aspect} 
& \textbf{Edge-based elicitation} 
& \textbf{Ordering-based elicitation} \\
\hline
Question format
& When the question is presented as $A$--$B$, please use $1$ to denote a direct causal influence from $A \rightarrow B$, use $-1$ to denote $A \leftarrow B$, and use $0$ to denote no direct influence between $A$ and $B$.
& How strongly do you believe that $A$ is an upstream causal variable of $B$? \\
\hline
Response space
& Integer score in $\{-1,0,1\}$
& Integer score in $[-10,10]$ \\
\hline
Direction encoding
& Encoded by the sign of the score
& Encoded by the sign of the score \\
\hline
Causal interpretation
& Direct causal influence only
& Upstream causal influence, including both direct and indirect effects \\
\hline
Example in the \textit{Asia} network
& For $(\text{Smoking}, \text{Lung Cancer})$, participants indicate whether they believe there is a direct causal edge.
& For $(\text{Smoking}, \text{Dyspnea})$, participants may give a positive score if they believe Smoking is upstream of Dyspnea through direct or indirect causal paths. \\
\hline
\end{tabular}}
\end{table}

\subsection{BN-Bench: Human and LLM Causal Knowledge Benchmark}
\label{bnbench}

To support the crowd-based causal learning setting, we develop BN-Bench, an evolving benchmark database for eliciting, aggregating, and evaluating causal knowledge from human participants and LLM-simulated participants. BN-Bench provides the data infrastructure needed for the ideas developed in the main paper. Each ground-truth network is represented as a set of directed causal chains, where each chain corresponds to an ordered causal path and the union of these chains defines the associated DAG. For each network, the database can store both human ratings and LLM-generated ratings under two elicitation protocols. The benchmark, including the current data organization and usage instructions, is available at
\url{https://anonymous.4open.science/r/BN-Bench-9D04/README.md}.

BN-Bench supports several functions that are central to crowd-based causal learning. It provides standardized network structures and pairwise causal ratings for evaluating causal graph recovery from noisy and heterogeneous judgments. It also supports comparisons between human participants and LLM-simulated participants, allowing users to study when LLMs approximate, complement, or differ from human causal beliefs. The benchmark further enables aggregation experiments in which individual judgments are combined into a collective causal structure and evaluated against the ground-truth DAG. Since the data are organized by network, source type, elicitation protocol, and rating file, BN-Bench can be used directly in reproducible Python workflows for loading networks, computing implied DAGs, aggregating ratings, and evaluating recovered graphs.

The current version of BN-Bench should be viewed as a growing database. Future releases will extend the collection to additional Bayesian networks, domains, elicitation protocols, expert populations, and LLM profiles. In this way, BN-Bench turns the paper's broader argument into a practical resource: crowd-based causal learning requires not only new algorithms, but also reusable datasets and infrastructure for collecting and comparing distributed causal knowledge in a systematic way.

\clearpage

\section{Stability of Crowd Aggregation under Random Participant Subsets}
\label{app:random_subsets}

We provide a preliminary empirical analysis of the scalability of crowd-based causal ordering. Following prior work \cite{xiao2018optimal}, we consider both the \textsc{Alarm} network with $37$ nodes and a randomly generated larger network with $100$ nodes. We simulate five experts with three quality levels, good, moderate, and poor, in a ratio of $1{:}3{:}1$, and collect pairwise comparisons under query budgets of $20$, $50$, and $100$. 

\begin{figure}[!h]
    \centering
    \includegraphics[width=0.6\textwidth]{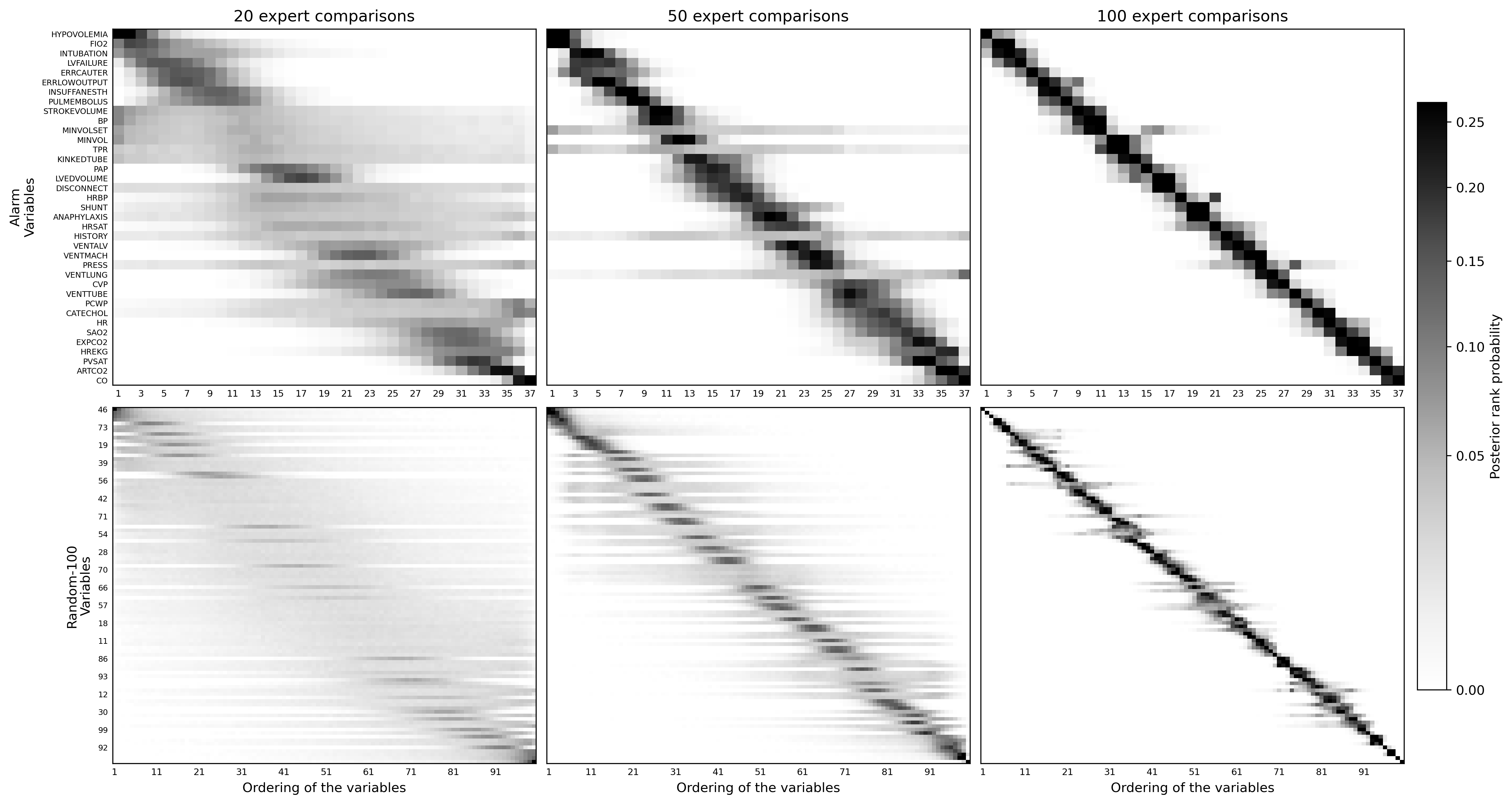}
    \caption{Uncertainty of the causal ordering under average aggregation of expert comparisons from five experts (ordered by accuracy from high to low: one good expert, three moderate-quality experts, and one poor expert) across two networks of different sizes. The top row shows results for the Alarm network ($N=37$), and the bottom row shows results for the Random-100 network ($N=100$). From left to right, the panels correspond to 20, 50, and 100 expert comparisons, respectively. Darker shades indicate higher posterior rank probability. The rows correspond to variables, and the x-axis indicates the ordering positions of the variables.}
    \label{fig:uncertainty_expert}
\end{figure}

\begin{figure}[!t]
    \centering
    \includegraphics[width=0.6\textwidth]{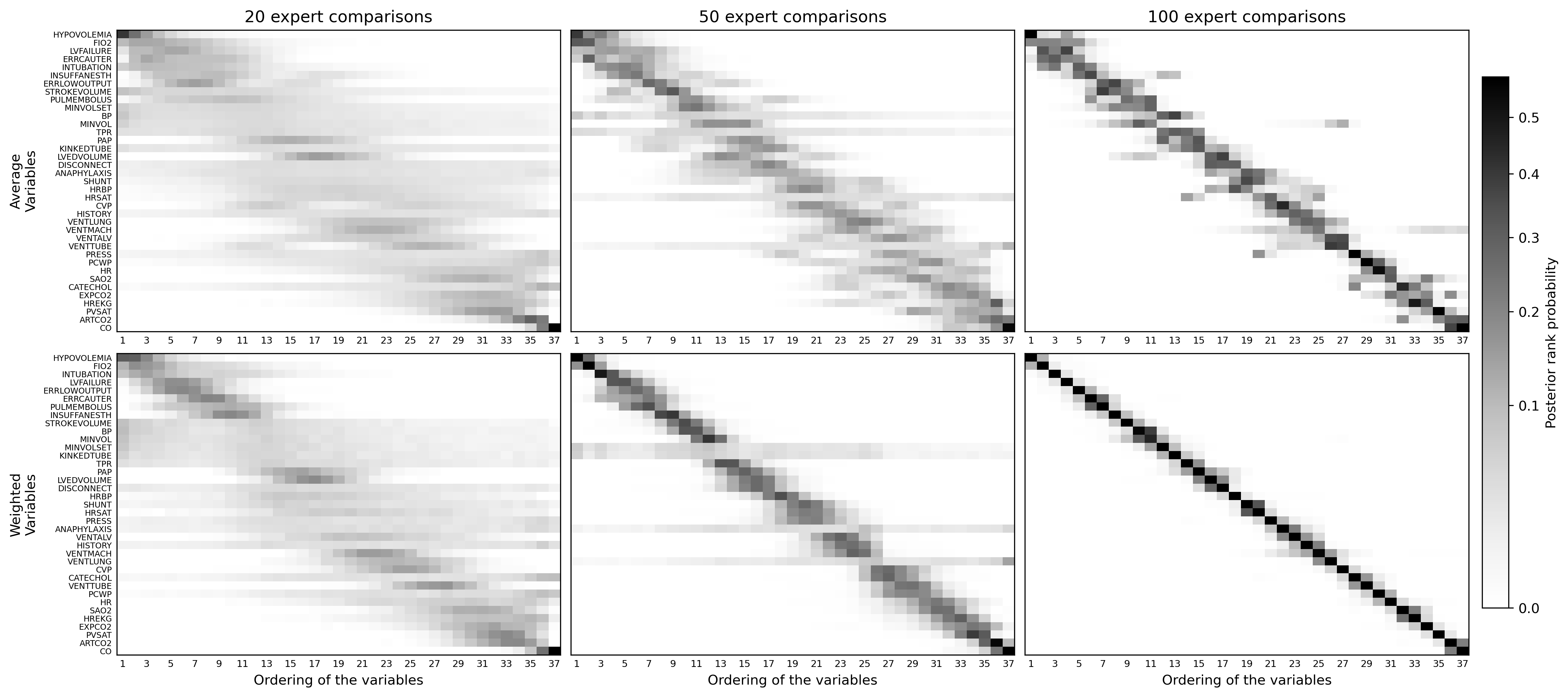}
    \caption{Uncertainty of the causal ordering under two aggregation strategies for expert comparisons from five experts (ordered by accuracy from high to low: one good expert, three moderate-quality experts, and one poor expert) in the Alarm network ($N=37$). The top row shows results under simple average aggregation, whereas the bottom row shows results under inverse-variance-weighted aggregation, which assigns larger weight to higher-accuracy experts and smaller weight to lower-accuracy experts. From left to right, the panels correspond to 20, 50, and 100 expert comparisons, respectively. Darker shades indicate higher posterior rank probability. The rows correspond to variables, and the x-axis indicates the ordering positions of the variables.}
    \label{fig:ave_vs_weight}
\end{figure}

As shown in Figure~\ref{fig:uncertainty_expert}, even simple averaging produces a visible diagonal alignment between the aggregated pairwise signal and the ground-truth causal order, suggesting that the crowd signal remains informative for ordering recovery. The larger $100$-node network converges more slowly under a very limited budget, especially with only $20$ comparisons, but increasing the budget to $100$ leads to a clearer concentration around the true order. We also compare simple averaging with a structured aggregation strategy on the \textsc{Alarm} network in Figure~\ref{fig:ave_vs_weight}. Structured aggregation recovers the true causal order more clearly and with smaller uncertainty across query budgets, indicating that aggregation design is an important component of crowd-based causal learning. 

These results provide preliminary evidence that pairwise comparisons under crowd aggregation can scale to larger networks, while a rigorous sample-complexity analysis remains an important direction for future work.

\clearpage

\section{Systematic Assessment: Edge-Wise v.s Ordering-Wise }
The following table assesses the two elicitation frameworks (edge-wise and ordering-wise) along four expert characteristics for understanding their relationships.
\begin{table*}[!h]
\centering
\caption{Comparison of edge-wise and ordering-wise causal knowledge. ($\uparrow$) / ($\downarrow$) denote performance improvement / degradation.}
\label{tab:expertType_edge_order}
\renewcommand{\arraystretch}{1.15}
\resizebox{\linewidth}{!}{
\begin{tabular}{l|l|l|l}
\toprule
\textbf{Characteristic} & \textbf{Edge-wise} & \textbf{Ordering-wise} & \textbf{Useful Metrics}\\
\midrule
\multirow{2}{*}{Completeness} & Incompleteness leads to missing & Allow sparse local coverage &  Edge coverage, recall\\
& (More edge information per query $\uparrow$) & (Higher pruning per query $\uparrow$) & Path coverage, path recall\\ 
\hline
\multirow{2}{*}{Belief validity} &  Local error, harder to detect & Error propagate structurally, easier to detect & Edge precision, FDR \\
& (Cumulate more errors v.s more edge info, $\downarrow\uparrow$) & (More info for pattern detection $\uparrow$) & Pairwise order accuracy, rank-based correlation\\ 
\hline
\multirow{2}{*}{Confidence level} & Not enough info, need repeated queries & Soft constraints, reveal belief geometry & Abstention rate, re-query variance for one edge\\
& (- if no repeated, query familiar $\uparrow$, o.w.,$\downarrow$)& (Query familiar $\uparrow$, o.w., $\downarrow$) & Score dispersion, posterior over orders \\ 
\hline
\multirow{2}{*}{Trustworthiness} & Direct contaminate structure & Global contradiction, enable screening  & Cycle injection rate, edge flip frequency \\
& (More reliable info $\uparrow$) & (More reliable responses or more info $\uparrow$) & Inconsistency graph count, rank-cycle frequency  \\ 
\bottomrule
\end{tabular}}
\end{table*}

\section{Causal Inference with Human Knowledge}\label{causal_inf}
Causal inference, a major branch of causal learning different from causal discovery, dominates applications such as recommendation systems \cite{wang2022causal, luo2024survey}, clinical trials \cite{piantadosi2024clinical}, and policy making \cite{imbens2024causal}. Identifying interventional effect from observational data typically requires stronger assumptions \cite{ding2018causal}. Recently, incorporating human knowledge into causal inference has emerged as a promising way to facilitate synergy with statistical learning.

To show the advantage of using human knowledge, we consider a simplified tripartite causal structure with $(N - 2)$ instrumental variables~(IVs) $\mathbf{X}_{IV} = (X_1, \ldots, X_{N-2})$ influencing a single exposure $\mathbf{X}_{E} = X_{N-1}$, and further affecting an outcome $\mathbf{X}_O = X_N$, i.e., $Pa_G(X_O) \subset \mathbf{X}_E$ and $Pa_G(X_E) \subset \mathbf{X}_{IV}$, where $Pa_G(\cdot)$ denotes the set of parent variables in $G$. IV methods offer a principled strategy to remove bias from unobserved confounding by isolating exogenous variation in treatment \cite{angrist1996identification}. With linearity assumptions, it can be cast as a two-stage model:  $X_E  =\mathbf{X}_{IV}\boldsymbol{\alpha} + U\xi + \epsilon_E$, $ X_O = X_E\beta + \mathbf{X}_{IV}\boldsymbol{\gamma} + U\zeta + \epsilon_o. $
Causal inference relies on the accurate estimation of $\beta$ and its corresponding hypothesis testing. Valid IVs must satisfy three important assumptions: 
\begin{assumption}[Valid instrumental variables (IVs)]
Valid IVs must satisfy three important assumptions: 
\begin{enumerate}
    \item \textit{Relevance:} IVs are strongly correlated with exposure; 
    \item \textit{Exclusion restriction:} IVs cannot have a direct effect on the outcome, nor can they affect the outcome through any path other than the treatment; 
    \item \textit{Exchangeability:} IVs must not be associated with any confounders (measured or unmeasured) that affect the outcome. 
\end{enumerate}  
\end{assumption}

These assumptions are critical for proper causal inference. Ensuring IVs meet these assumptions has been an active research area, as weak or invalid IVs will lead to biased and then incorrect testing results \cite{wu2025instrumental, guo2023causal, lin2024instrumental,kang2025identification}. More specifically, the ideal case where we know the correct set of strong IVs $\mathbf{X}_{IV,s}$, $\hat{\beta} = \left(\hat{\mathbf{X}}_E^{T}\hat{\mathbf{X}}_E\right)^{-1}\hat{\mathbf{X}}_E^T\mathbf{X}_O, \hat{\mathbf{X}}_E = \mathbf{X}_{IV, s}\left(\mathbf{X}_{IV,s}^T\mathbf{X}_{IV,s}\right)^{-1}\mathbf{X}_{IV,s}^T\mathbf{X}_E$. The inclusion of the other IVs results in an additional term as $\beta + e_I$, where $e_I$ depends on both the correlation among IVs and their respective effect sizes on exposure and outcome, potentially overwhelming the true causal effect \cite{burgess2017review, burgess2021mendelian}. Incorporation of human knowledge will be greatly beneficial if it can tell some IVs should be excluded from the model directly and even more helpful when the validation of structure and assumptions gets more challenging in nonlinear or weak signal scenarios.   
\section{Examples of Prompts for LLM Experts}

We provide example prompt templates used to elicit causal judgments from LLM-based simulated experts. The first template asks the LLM expert to rate the causal ordering of each queried variable pair on a signed integer scale. The second template performs a self-verification step, in which the LLM revisits its initial judgmentss, and revises uncertain ratings when appropriate. 
The details are summarized in Tables~\ref{tab:llm_prompt_example1} and~\ref{tab:llm_prompt_example2}.

\begin{table*}[!htbp]
\centering
\caption{Prompt template for eliciting causal judgments from LLM experts.}
\label{tab:llm_prompt_example1}

\begin{tcolorbox}[
    width=\textwidth,
    colback=blue!4!white,
    colframe=cyan!45!black,
    coltitle=white,
    colbacktitle=cyan!45!black,
    title={\textsc{Elicit Causal Judgments from LLM (Single-Step)}},
    fonttitle=\bfseries\large,
    boxrule=0.8pt,
    arc=2mm,
    left=10pt,
    right=10pt,
    top=8pt,
    bottom=8pt
]
\begin{quote}
You are an expert in \texttt{[Background Clarification]}. 
Based on your domain knowledge and the variable definitions provided below, please answer the following causal judgment survey.

\medskip
\textbf{Background.} 
We consider the following variables:
\[
\texttt{[Variable list with short definitions]}
\]
Your task is to assess whether one variable is causally upstream of another variable. 
Here, ``Factor A is an upstream causal variable of Factor B'' means that changes in Factor A may directly or indirectly influence Factor B through a plausible causal pathway. 

\medskip
For each queried pair $(A,B)$, please answer the question: ``How strongly do you believe that Factor A is an upstream causal variable of Factor B $(A \rightarrow B)$?" Please use an integer rating from $-10$ to $10$.

\medskip
The rating scale ranges from $-10$ to $10$: $+10$ indicates strong evidence that $A$ is upstream of $B$, $+5$ indicates moderate evidence that $A$ is upstream of $B$, $0$ indicates no clear causal ordering or no perceived causal relationship, $-5$ indicates moderate evidence that $B$ is upstream of $A$, and $-10$ indicates strong evidence that $B$ is upstream of $A$.

\medskip
\textbf{Input} A list of queried variable pairs: \texttt{[Pair list]}

\medskip
% \textbf{Output format}
% Return one line for each pair in the following format:
% \[
% (A,B): \text{rating}; \text{brief justification}
% \]

\textbf{Output:} \texttt{[Rating answers]}
\end{quote}
\end{tcolorbox}
\end{table*}

\begin{table*}[!htbp]
\centering
\caption{Prompt template for self-verification of the elicited judgments.}
\label{tab:llm_prompt_example2}

\begin{tcolorbox}[
    width=\textwidth,
    colback=blue!4!white,
    colframe=cyan!45!black,
    coltitle=white,
    colbacktitle=cyan!45!black,
    title={\textsc{Elicit Causal Judgments from LLM (Verification)}},
    fonttitle=\bfseries\large,
    boxrule=0.8pt,
    arc=2mm,
    left=10pt,
    right=10pt,
    top=8pt,
    bottom=8pt
]
\begin{quote}
You are given the causal ratings produced in the previous step:

\[
\texttt{[Initial causal ratings and justifications]}
\]

Please verify these judgments carefully using the same background knowledge and variable definitions:
\[
\texttt{[Background Clarification and Variable Definitions]}
\]

After verification, provide the final rating using the same integer scale from $-10$ to $10$, where positive values indicate that $A$ is upstream of $B$, negative indicate that $B$ is upstream of $A$, and $0$ indicates no clear causal ordering.

\medskip
% \textbf{Output format}
% For each pair, return:
% \[
% (A,B): \text{initial rating} \rightarrow \text{final rating}; \text{brief reason for confirmation or revision}
% \]

\textbf{Output:} \texttt{[Verified rating answers]}.
\end{quote}
\end{tcolorbox}
\end{table*}

\section{Topics for Future Research}\label{future}
Please see Table \ref{tab:future_directions}. Here, we only outline a limited set of research questions, meant to be illustrative rather than exhaustive. We leave ample room for more open questions to be explored in the future.
\begin{table*}[!h]
\centering
\small
\caption{Future research directions for crowd-powered causal learning, categorized by modeling, elicitation, human–AI collaboration, and broader systems challenges. Each challenge includes specific questions to advance this emerging field.}
\label{tab:future_directions}
% \small
\resizebox{\linewidth}{!}{
\begin{tabular}{|p{2cm}|p{2.5cm}|p{11cm}|}
\hline
\textbf{Category} & \textbf{Research \newline Challenge} & \textbf{Key Questions} \\
\hline

& \vspace{2em}Characterizing the knowledge profile of an expert &
\begin{itemize}[itemsep=0pt, leftmargin=*]
  \item Can we build a unified statistical model, i.e., by probabilistic graphical models, that can universally characterize human causal knowledge adequately?
  \item Can topic or embedding methods capture an expert’s domain familiarity?
  \item How can we detect and represent structured expertise, such as an expert’s strength in local subnetworks but not globally?
  \item Can we build interpretable reliability maps across the graph for each expert?
\end{itemize}  \\ \cline{2-3}

\vspace{1em}Expert \newline Knowledge \newline Modeling& \vspace{1em} Modeling \newline structured \newline uncertainty or \newline abstention & 
\begin{itemize}[itemsep=0pt, leftmargin=*]
  \item How can we model expert uncertainty in a structured way (e.g., confidence scores, conditional statements)?
  \item Can abstentions be modeled as informative signals rather than nulls?
  \item What are formal representations of ``partial knowledge'' beyond edge judgments (e.g., subgraph templates, logical rules)?
\end{itemize} \\ \cline{2-3}

& \vspace{1em} Fusing \newline heterogeneous \newline experts & 
\begin{itemize}[itemsep=0pt, leftmargin=*]
  \item How to integrate mixed forms of input: edges, orderings, counterfactuals?
  \item Can we define a unifying representation that respects all input formats?
  \item How to resolve conflicting inputs across representation types and reliability levels?
  \item Can we detect and screen out bad actors to avoid malicious behaviors?
\end{itemize}\\ \cline{2-3}

& \vspace{1em} Adapting expert \newline modeling over \newline time & 
\begin{itemize}[itemsep=0pt, leftmargin=*]
  \item Can we update beliefs about expert accuracy online as data come in?
  \item What signals should we track to detect changes in knowledge quality?
  \item Can interactive systems personalize which graph regions to route to each expert?
\end{itemize}\\

\hline

& \vspace{1em} Optimal \newline elicitation \newline strategies under  \newline a fixed budget & 
\begin{itemize}[itemsep=0pt, leftmargin=*]
  \item What strategies maximize expected information gain per question?
  \item Can we prioritize resolving parts of the graph with the highest causal ambiguity?
  \item Can budget-aware strategies be adapted dynamically as expert responses arrive?
\end{itemize} \\ \cline{2-3}

&\vspace{1em}  Selecting between query types & 
\begin{itemize}[itemsep=0pt, leftmargin=*]
  \item When should we ask about edges vs. causal orderings vs. counterfactuals?
  \item Can we learn which query types are more accurate for different experts?
  \item How to model and exploit the costs and uncertainties of different query types?
\end{itemize} \\ \cline{2-3}

Elicitation \newline Optimization& \vspace{2em} Identifying the \newline right respondent & 
\begin{itemize}[itemsep=0pt, leftmargin=*]
  \item Can we predict expert suitability using metadata (e.g., profession, past accuracy, self-reported expertise)?
  \item How can we personalize query routing to match each contributor’s strengths?
  \item Can learning-to-route models optimize matching of questions to experts adaptively?
\end{itemize} \\ \cline{2-3}

& \vspace{0.5em} Interpretable and  \newline engaging \newline elicitation & 
\begin{itemize}[itemsep=0pt, leftmargin=*]
  \item What interfaces help non-experts express causal knowledge accurately?
  \item Can we visualize uncertainty or partial graphs to guide intuitive judgments?
  \item Can confidence calibration tools help users assess when to abstain?
\end{itemize} \\
\hline

\end{tabular}}
% \caption{Future research directions for crowd-powered causal learning, categorized by modeling, elicitation, human–AI collaboration, and broader systems challenges. Each challenge includes specific questions to advance this emerging field.}
 \label{tab:future_directions}
\end{table*}

\clearpage 
\begin{table*}[!t]
\centering
\small
\resizebox{\linewidth}{!}{
\begin{tabular}{|p{2cm}|p{2.5cm}|p{11cm}|}
\hline
\textbf{Category} & \textbf{Research \newline Challenge} & \textbf{Key Questions} \\
\hline

& \vspace{2em} Optimizing \newline division of labor & 
\begin{itemize}[itemsep=0pt, leftmargin=*]
  \item When should a query be routed to an LLM vs. a human?
  \item How to dynamically reassign tasks based on model confidence or experts' quality?
  \item How to combine LLM reasoning and crowd input into unified learning pipelines?
\end{itemize} \\ \cline{2-3}

\vspace{-2em}Human–AI \newline Collaboration&\vspace{0.5em}  LLMs as \newline elicitation \newline intermediaries & 
\begin{itemize}[itemsep=0pt, leftmargin=*]
\item Can LLMs help translate between formal causal concepts and human-friendly prompts, or refine user input into structured formats?
\item If so, how to model this process, and do quality control and optimization?
\end{itemize}
 \\ 
 
 % \cline{2-3}

% & LLMs simulating expert priors & TBD \\ \cline{2-3}

% & Validating and  \newline refining LLM-generated structures & TBD \\ 

\hline
& \vspace{2.5em} Fundamental \newline limits of \newline crowd-based \newline causal learning & 
\begin{itemize}[itemsep=0pt, leftmargin=*]
  \item Identifiability limits when human knowledge is partial, inconsistent, or noisy?
  \item Can we derive information-theoretic lower bounds on the number or type of queries required to recover causal structure?
  \item Do graph properties (e.g., sparsity, treewidth) affect learnability from crowd input?
  \item What theoretical guarantees can we provide when combining observational data with distributed human knowledge?
\end{itemize} \\ \cline{2-3}

\vspace{2em}Broader \newline  Theoretical and  \newline Systems \newline Questions
 & \vspace{4em} Benchmark \newline development & 
\begin{itemize}[itemsep=0pt, leftmargin=*]
  \item What benchmark tasks and datasets are needed to evaluate systems that use human input for causal discovery?
  \item How can we simulate crowd-like behaviors (e.g., imperfect expertise, varying reliability) to support controlled comparisons?
  \item What metrics capture the quality of learned causal structures when ground truth is incomplete or unavailable?
  \item How should we evaluate human effort efficiency and knowledge acquisition cost in interactive systems?
\end{itemize} \\ \cline{2-3}

& \vspace{2.5em} Ethical and  \newline equitable \newline participation & 
\begin{itemize}[itemsep=0pt, leftmargin=*]
  \item How to ensure inclusivity and reduce bias in knowledge collection?
  \item What incentives encourage high-quality participation from diverse contributors?
  \item How to ensure transparency, attribution, and accountability in aggregating crowd knowledge?
  \item What risks arise when aggregating causal beliefs across cultural, disciplinary, or epistemic divides?
\end{itemize} \\

\hline
\end{tabular}}
\end{table*}
\end{appendices}
\end{document}